%% file: main.tex
\documentclass{article}
\usepackage[preprint]{colm2026_conference}

\usepackage{microtype}
\usepackage{hyperref}
\usepackage{url}
\usepackage{booktabs}
\usepackage{array}
\usepackage{graphicx}
\usepackage{lineno}
\usepackage{multirow}
\usepackage{xcolor}
\usepackage{colortbl}
\usepackage{enumitem}
\usepackage{tikz}
\usetikzlibrary{positioning,arrows.meta,fit,backgrounds,calc}

\input{math_commands}

\definecolor{darkblue}{rgb}{0, 0, 0.5}
\hypersetup{colorlinks=true, citecolor=darkblue, linkcolor=darkblue, urlcolor=darkblue}

\title{Structured Distillation of Web Agent Capabilities\\Enables Generalization}

\author{
Xing Han L\`{u} \\
Mila -- Quebec AI Institute \\
McGill University \\
\texttt{xing.han.lu@mail.mcgill.ca}
\And
Siva Reddy \\
Mila -- Quebec AI Institute \\
McGill University
}

\newcommand{\ours}{\textsc{Agent-as-Annotators}}

\newcommand{\websynth}{\textsc{A3-Synth}}

\begin{document}

\ifcolmsubmission
\linenumbers
\fi

\maketitle

\begin{abstract}
Frontier LLMs can navigate complex websites, but their cost and reliance on third-party APIs make local deployment impractical. We introduce \ours{}, a framework that structures synthetic trajectory generation for web agents by analogy to human annotation roles, replacing the Task Designer, Annotator, and Supervisor with modular LLM components. Using Gemini~3~Pro as teacher, we generate 3{,}000 trajectories across six web environments and fine-tune a 9B-parameter student with pure supervised learning on the 2{,}322 that pass quality filtering. The resulting model achieves 41.5\% on WebArena, surpassing closed-source models such as Claude~3.5~Sonnet (36.0\%) and GPT-4o (31.5\%) under the same evaluation protocol, and nearly doubling the previous best open-weight result (Go-Browse, 21.7\%). Capabilities transfer to unseen environments, with an 18.2 percentage point gain on WorkArena~L1 (an enterprise platform never seen during training) and consistent improvements across three additional benchmarks. Ablations confirm that each pipeline component contributes meaningfully, with Judge filtering, evaluation hints, and reasoning traces each accounting for measurable gains. These results demonstrate that structured trajectory synthesis from a single frontier teacher is sufficient to produce competitive, locally deployable web agents. Project page: \url{https://agent-as-annotators.github.io}
\end{abstract}

\section{Introduction}
\label{sec:intro}

Frontier LLMs can now complete realistic web tasks~\citep{Wang2023ASO, Yao2022ReActSR, Wei2022ChainOT}, from filling out forms and querying databases to managing content across multiple applications. On WebArena~\citep{Zhou2023WebArenaAR}, a benchmark of self-hosted web applications, multi-agent systems exceed 70\% success rate on public leaderboards~\citep{Guo2026OpAgentOA}, and even single frontier models reach over 50\% under standardized evaluation (BrowserGym;~\citealp{Chezelles2024TheBE}). However, these models require expensive API access, transmit user data to third-party servers, and cannot be run locally. Small open-weight models (${\sim}$9B parameters)~\citep{Yang2024Qwen25TR} offer an attractive alternative but exhibit a substantial capability gap, trailing frontier models by over 22 percentage points on WebArena.

A natural approach to closing this gap is \emph{agentic capability distillation}: using a frontier model as a teacher to generate training trajectories for a smaller student~\citep{Hsieh2023DistillingSO}. Several recent works explore this for web agents: InSTA~\citep{Trabucco2025TowardsIT} proposes tasks at internet scale across 150K websites, and NNetNav~\citep{Murty2024NNetNavUL} retroactively labels exploration trajectories with task descriptions. These pipelines are effective, but their designs differ in ways that are difficult to compare systematically. What is needed is a common framework for describing the roles that each pipeline component plays.

\begin{figure}[t]
\centering
\includegraphics[width=\linewidth]{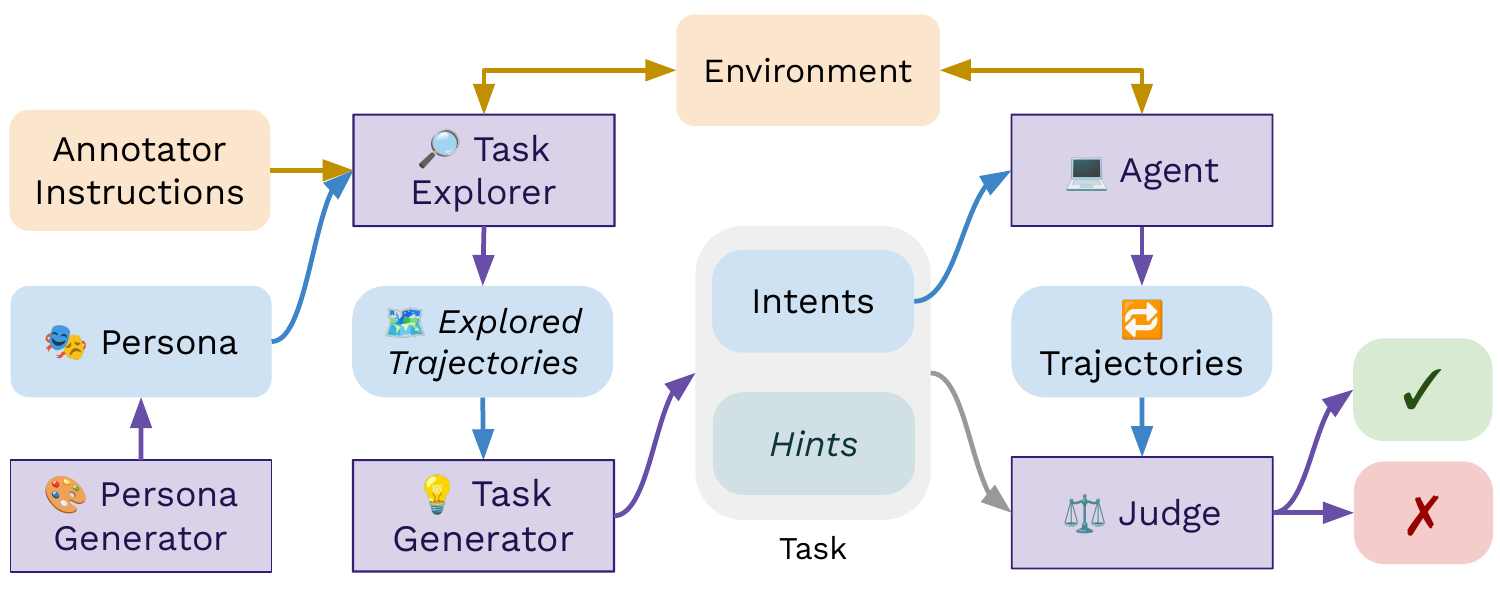}
\caption{The \ours{} pipeline replaces three human annotation roles with LLM modules. The Task Designer is replaced by a \textbf{Persona Generator} and \textbf{Task Generator} that synthesize task intents with evaluation hints; the Annotator is replaced by an \textbf{Agent}; the Supervisor is replaced by a \textbf{Judge}. Only successful trajectories train the student.}
\label{fig:framework}
\end{figure}

We propose \ours{}, a framework inspired by the established practice of \emph{human annotation} for web agent benchmarks. When creating WebArena's evaluation tasks, human contributors played three distinct roles: a \textbf{Task Designer} who explored the environment and designed tasks with evaluation hints, an \textbf{Annotator} who executed tasks to produce trajectories, and a \textbf{Supervisor} who verified completion. \ours{} replaces each role with LLM modules: a Persona Generator and Task Generator (Task Designer), an Agent (Annotator), and a Judge with evaluation hints (Supervisor), as shown in \figref{fig:framework}. Prior pipelines such as InSTA and NNetNav are naturally expressed as instantiations of this framework with specific module choices (Table~\ref{tab:framework_comparison}).

We implement \ours{} using Gemini~3~Pro as the teacher to generate \websynth{}, a synthesized training set of 3,000 tasks across six WebArena environments (16,353 training examples after filtering). A 9B student fine-tuned on this data achieves 41.5\% on WebArena, surpassing GPT-4o and Claude~3.5~Sonnet under the same evaluation protocol (Table~\ref{tab:cross_benchmark}), and transfers to completely unseen platforms including an 18.2pp gain on the enterprise WorkArena~L1 benchmark. Our contributions are: (1)~\ours{}, a framework that organizes trajectory synthesis pipelines by analogy to human annotation roles, enabling systematic comparison (\secref{sec:framework}); (2)~\websynth{}, a synthesized dataset of 3,000 web tasks with evaluation hints (\secref{sec:implementation}); (3)~empirical findings that teacher quality matters more than data quantity, lower reasoning budgets produce better training data, and web skills transfer broadly to unseen environments (\secref{sec:results}).

\section{Related Work}
\label{sec:related}

\paragraph{Web agent benchmarks and architectures.}
Web agent evaluation has progressed from synthetic micro-environments such as World of Bits~\citep{Shi2017WorldOB} and MiniWoB~\citep{Liu2018ReinforcementLO} through simulated e-commerce~\citep{Yao2022WebShopTS} and cross-website generalization~\citep{Deng2023Mind2WebTA, Lu2024WebLINXRW} to realistic self-hosted environments. WebArena~\citep{Zhou2023WebArenaAR} deploys six web applications with 812 tasks evaluated by functional correctness; VisualWebArena~\citep{Koh2024VisualWebArenaEM} extends this to vision-grounded tasks; WorkArena~\citep{Drouin2024WorkArenaHC} and WorkArena++~\citep{Boisvert2024WorkArenaTC} target enterprise interfaces; OSWorld~\citep{Xie2024OSWorldBM} and AndroidWorld~\citep{Rawles2024AndroidWorldAD} broaden to desktop and mobile. BrowserGym~\citep{Chezelles2024TheBE} unifies many of these under a common interface, which we use for both data generation and evaluation. For a comprehensive survey, see~\citet{Wang2024GUIAW}. Notably, the task creation process for these benchmarks, involving task design, annotation, and supervision roles, directly inspires our framework.

\paragraph{Synthetic trajectory generation for web agents.}
The high cost of human demonstrations has motivated LLM-based trajectory synthesis. InSTA~\citep{Trabucco2025TowardsIT} proposes and executes tasks at internet scale across 150K websites, filtering with an LLM judge. NNetNav~\citep{Murty2024NNetNavUL} retroactively labels free exploration trajectories with task descriptions and filters via an outcome reward model. AgentTrek~\citep{Xu2024AgentTrekAT} leverages web tutorials for task replay with VLM verification. Explorer~\citep{Pahuja2025ExplorerSE} scales exploration-driven synthesis to 94K trajectories by iteratively refining exploration into task descriptions. Go-Browse~\citep{Gandhi2025GoBrowseTW} frames collection as a graph search over URLs with VLM feasibility checking. These approaches are effective but differ in pipeline design; Table~\ref{tab:framework_comparison} maps them onto \ours{} modules, and we discuss the key structural differences in \secref{sec:prior_work}. More broadly, FireAct~\citep{Chen2023FireActTL}, AgentTuning~\citep{Zeng2023AgentTuningEG}, and WebRL~\citep{Qi2024WebRLTL} explore agent fine-tuning on LLM-generated trajectories.

\paragraph{Knowledge distillation and synthetic data.}
Using stronger models to train weaker ones has proven effective across many settings, from Self-Instruct~\citep{Wang2022SelfInstructAL} and persona-driven diversity~\citep{Chan2024ScalingSD} to reasoning distillation~\citep{Mukherjee2023OrcaPL, Hsieh2023DistillingSO} and curated synthetic data~\citep{Gunasekar2023TextbooksAA}. LIMA~\citep{Zhou2023LIMALI} showed that data quality compensates for quantity. \citet{Penaloza2026PrivilegedID} study distillation with privileged information, where the teacher accesses signals unavailable to the student at inference; our pipeline follows a similar principle, as the teacher's exploration data and evaluation hints inform trajectory generation but are not provided to the student. We extend these insights to the agentic setting, where training data consists of multi-step environment interaction trajectories.

\paragraph{LLM-based evaluation and self-improvement.}
LLMs as evaluators~\citep{Zheng2023JudgingLW, Kim2023PrometheusIF} are now widespread; AgentRewardBench~\citep{lu2025agentrewardbench} provides a meta-evaluation benchmark for assessing such automatic evaluators in the web agent setting. Our Judge augments the LLM-as-judge approach with evaluation hints for improved reliability. Complementary RL-based approaches, including DigiRL~\citep{Bai2024DigiRLTI}, Agent~Q~\citep{Putta2024AgentQA}, and OpenWebVoyager~\citep{He2024OpenWebVoyagerBM}, could further refine agents produced by our SFT pipeline.

\begin{table}[t]
\centering
\caption{Distillation gains transfer across all five benchmarks, with the largest improvement on WorkArena~L1 (+18.2pp), an enterprise interface never seen during training. Success rate (\%); WA = WebArena, VWA = VisualWebArena, WoA = WorkArena~\citep{Drouin2024WorkArenaHC}, W++ = WorkArena++~\citep{Boisvert2024WorkArenaTC}, WoB = MiniWoB. \textit{$^*$Teacher model used to generate \websynth{}.}}
\label{tab:cross_benchmark}
\small
\begin{tabular}{lccccc}
\toprule
\textbf{Model} & \textbf{WA} & \textbf{VWA} & \textbf{WoA L1} & \textbf{W++ L2} & \textbf{WoB} \\
\midrule
\multicolumn{6}{l}{\textit{Proprietary}} \\
Gemini 3 Pro$^*$ & 51.2 & 49.0 & 79.7 & 41.6 & 74.7 \\
Gemini 3.1 Flash L. & 42.3 & 35.0 & 58.5 & 21.1 & 74.1 \\
\midrule
\multicolumn{6}{l}{\textit{Open-weight (base)}} \\
Qwen3.5-27B & 41.5 & 37.4 & 57.0 & 18.9 & 70.9 \\
Qwen3.5-9B & 31.0 & 28.5 & 33.3 & 2.2 & 63.2 \\
Qwen3.5-4B & 24.1 & 24.7 & 33.6 & 1.6 & 61.1 \\
Qwen3.5-2B & 3.1 & 5.3 & 4.2 & 0.0 & 11.8 \\
\midrule
\multicolumn{6}{l}{\textit{A3 fine-tuned (ours; teacher: Gemini 3 Pro)}} \\
\textbf{A3-Qwen3.5-9B} & \textbf{41.5} {\scriptsize\textcolor{teal}{(+10.5)}} & \textbf{33.9} {\scriptsize\textcolor{teal}{(+5.4)}} & \textbf{51.5} {\scriptsize\textcolor{teal}{(+18.2)}} & \textbf{9.7} {\scriptsize\textcolor{teal}{(+7.5)}} & \textbf{69.0} {\scriptsize\textcolor{teal}{(+5.8)}} \\
A3-Qwen3.5-4B & 35.2 {\scriptsize\textcolor{teal}{(+11.1)}} & 30.1 {\scriptsize\textcolor{teal}{(+5.4)}} & 44.8 {\scriptsize\textcolor{teal}{(+11.2)}} & 3.8 {\scriptsize\textcolor{teal}{(+2.2)}} & 66.9 {\scriptsize\textcolor{teal}{(+5.8)}} \\
A3-Qwen3.5-2B & 9.2 {\scriptsize\textcolor{teal}{(+6.1)}} & 7.6 {\scriptsize\textcolor{teal}{(+2.2)}} & 6.7 {\scriptsize\textcolor{teal}{(+2.5)}} & 0.0 & 38.6 {\scriptsize\textcolor{teal}{(+26.8)}} \\
\bottomrule
\end{tabular}
\end{table}

\section{The \ours{} Framework}
\label{sec:framework}

We describe how \ours{} maps human annotation roles to LLM modules (\secref{sec:human_annotation}), then show how prior trajectory synthesis pipelines are naturally expressed as instantiations (\secref{sec:prior_work}).

\subsection{From Human Roles to LLM Modules}
\label{sec:human_annotation}
\label{sec:llm_modules}

The creation of web agent training and evaluation data~\citep{Zhou2023WebArenaAR, Koh2024VisualWebArenaEM, Deng2023Mind2WebTA, Lu2024WebLINXRW} follows a structured human annotation process similar to those used in other NLP data collection efforts~\citep{Snow2008CheapAF}, where annotation quality and consistency are known challenges~\citep{Gururangan2018AnnotationAI}. We observe that this process involves three functional roles, present in varying degrees across benchmarks. A \textbf{Task Designer} explores the web environment, adopts a specific perspective, and produces tasks consisting of a natural language intent paired with evaluation criteria. In benchmarks that collect human demonstrations, such as Mind2Web~\citep{Deng2023Mind2WebTA} and WebLINX~\citep{Lu2024WebLINXRW}, an \textbf{Annotator} then receives the task intent and executes it on the environment, producing a step-by-step interaction trajectory. In benchmarks focused on evaluation, such as WebArena~\citep{Zhou2023WebArenaAR}, the Task Designer instead writes programmatic evaluation functions and the Annotator role is implicit (the agent under evaluation plays this role at test time). In both cases, a \textbf{Supervisor} reviews the results to verify quality, whether through manual inspection of trajectories or through evaluation criteria that check task completion.

\ours{} replaces each human role with LLM modules, organized into two phases that mirror the division of labor in human annotation.

\paragraph{Phase 1: Task synthesis (replacing the Task Designer).}
The \textbf{Persona Generator} produces diverse user personas (background, expertise, goals) that induce varied task distributions across different usage patterns within the same environment. The \textbf{Task Generator} receives a persona, annotator instructions, and access to the web environment. It explores the environment, stores observations, and synthesizes task intents paired with evaluation hints grounded in actual environment state. This grounding is critical: the Task Generator references real entities (existing users, products, repositories) rather than hallucinated ones.

\paragraph{Phase 2: Trajectory collection and filtering (replacing Annotator and Supervisor).}
The \textbf{Agent} receives only the task intent and interacts with a freshly reset environment to produce a trajectory, without access to hints, exploration data, or the persona. This separation ensures trajectories reflect genuine task-solving behavior. The \textbf{Judge} evaluates each trajectory using the interaction record \emph{and} the evaluation hints, which provide structured criteria that improve success assessment reliability, particularly for ambiguous final states. Only successful trajectories are retained for student training (Figure~\ref{fig:framework}).

\subsection{Prior Work as Framework Instantiations}
\label{sec:prior_work}

\begin{table}[t]
\centering
\caption{Trajectory synthesis pipelines mapped onto the \ours{} framework. \ours{} is the only pipeline that instantiates all six modules. Task: \textit{Ground.} = tasks defined before execution; \textit{Retro.} = labeled post-hoc. Shaded cells = module absent. Individual module contributions are ablated in Appendix~\ref{app:ablations}.}
\label{tab:framework_comparison}
\small
\setlength{\tabcolsep}{2pt}
\begin{tabular}{@{}lccccccl@{}}
\toprule
\textbf{Pipeline} & \textbf{Persona} & \textbf{Expl.} & \textbf{Task} & \textbf{Hints} & \textbf{Agent} & \textbf{Judge} & \textbf{Env.} \\
\midrule
NNetNav & Static DB & \checkmark & Retro. & \cellcolor{gray!15}-- & \cellcolor{gray!15}-- & LLM & WA+Web \\
InSTA & \cellcolor{gray!15}-- & \checkmark & Ground. & \cellcolor{gray!15}-- & \checkmark & LLM & Web \\
AgentTrek & \cellcolor{gray!15}-- & \cellcolor{gray!15}-- & Tutorial & \cellcolor{gray!15}-- & \checkmark & VLM & Web \\
Explorer & \cellcolor{gray!15}-- & \checkmark & Retro. & \cellcolor{gray!15}-- & \cellcolor{gray!15}-- & LLM & Web \\
Go-Browse & \cellcolor{gray!15}-- & \checkmark & Ground. & \cellcolor{gray!15}-- & \checkmark & VLM & WA \\
\midrule
\textbf{\ours{}} & \checkmark & \checkmark & Ground. & \checkmark & \checkmark & LLM & WA \\
\bottomrule
\end{tabular}
\end{table}

Table~\ref{tab:framework_comparison} maps five recent trajectory synthesis pipelines onto the \ours{} modules. A key structural distinction is \emph{task grounding}: whether tasks are defined before execution (\emph{grounded}) or retroactively extracted from exploration traces (\emph{retroactive}). Only grounded approaches can generate evaluation hints alongside intents, and only \ours{} currently exploits this.

\begin{itemize}[nosep,leftmargin=1.5em]
\item \textbf{InSTA}~\citep{Trabucco2025TowardsIT} (grounded): achieves diversity through scale (150K websites) rather than personas; uses an LLM judge without hints.
\item \textbf{Go-Browse}~\citep{Gandhi2025GoBrowseTW} (grounded): uses graph search over URLs with VLM feasibility checking and separate solver models.
\item \textbf{NNetNav}~\citep{Murty2024NNetNavUL} (retroactive): uses a fixed persona set; an LLM Explorer collects trajectories, then a Task Labeler assigns intents post-hoc.
\item \textbf{Explorer}~\citep{Pahuja2025ExplorerSE} (retroactive): fuses exploration, task generation, and execution into a single refinement loop at 94K-trajectory scale.
\end{itemize}

These design choices have downstream consequences; the framework makes them explicit and comparable (detailed vocabulary mappings in Appendix~\ref{app:vocab_mapping}). We focus on validating the complete pipeline and on teacher model quality, which we find to be the most important factor we studied (\secref{sec:teacher_quality}).

\section{Experimental Setup}
\label{sec:implementation}

We describe the \websynth{} dataset (\secref{sec:websynth}), the training procedure (\secref{sec:training}), and the evaluation benchmarks and baselines (\secref{sec:eval_setup}).

\subsection{\websynth{}: Synthesized Training Data}
\label{sec:websynth}

We implement \ours{} on the six self-hosted WebArena environments~\citep{Zhou2023WebArenaAR}: a Reddit forum, GitLab, an e-commerce site, its administration panel, Wikipedia, and OpenStreetMap. We generate 250 diverse personas, each assigned to all six environments for 1,500 explorations, and synthesize two task intents per exploration (from different steps), yielding 3,000 tasks in total (\websynth{}).

\paragraph{Teacher model.}
We use Gemini~3~Pro with a reduced thinking budget as the teacher model for both the Task Generator (exploration and task synthesis) and the Agent (trajectory collection). The same frontier model serves in both roles, simplifying the pipeline. We configure the model to produce concise reasoning traces rather than extended deliberation; counterintuitively, this reduced thinking budget achieves \emph{higher} success rates on \websynth{} tasks across all six environments than the default configuration, a finding we analyze in \secref{sec:teacher_quality}. We also evaluate alternative teachers (Gemini~3.1~Pro, Gemini~3~Flash) in Appendix~\ref{app:teacher_comparison}.

\paragraph{Judge.}
The Judge module is also implemented with an LLM (Gemini~3~Pro), which receives the agent's full interaction trajectory, the task intent, and the evaluation hints. The hints provide structured information about the expected outcome (e.g., ``the user should see a confirmation message on the settings page'' or ``the repository should contain a new file named X''). This design draws on the LLM-as-judge paradigm~\citep{Zheng2023JudgingLW} but augments it with task-specific hints that substantially aid in determining success for tasks where the final state is otherwise ambiguous. The Judge answers four standardized evaluation questions about the trajectory and produces a binary success/failure label; to mitigate position bias, we randomize the ordering of evaluation options.

\paragraph{Data statistics.}
After Judge filtering, the teacher produces successful trajectories for 69--85\% of tasks depending on the environment (Table~\ref{tab:websynth}), yielding 2,322 successful trajectories comprising 16,353 training examples (observation-action step pairs) with an average of 7.0 steps per trajectory and an average response length of 1,920 characters. All responses contain explicit reasoning traces in structured blocks (average 1,021 characters of reasoning per response). The data scaling curve (\secref{sec:ablations}) shows diminishing returns at this scale, suggesting that further gains may require broader environment coverage or complementary approaches rather than simply more trajectories from the same pipeline.

\subsection{Training}
\label{sec:training}

Successful trajectories are converted to multi-turn SFT format~\citep{Ouyang2022TrainingLM}. Each interaction step becomes one exchange: a user message containing the observation (accessibility tree + screenshot + task goal) and an assistant response containing structured reasoning (in \texttt{<thought>} and \texttt{<think>} blocks) followed by the action. We apply cross-entropy loss on assistant tokens only.

\paragraph{Student model and training.}
We fine-tune Qwen3.5-9B~\citep{Bai2025Qwen25VLTR}, a 9B multimodal model supporting text and image inputs, for 2 epochs (${\sim}$1,022 steps) using FSDP~\citep{Zhao2023PyTorchFE} with FlashAttention~\citep{Dao2022FlashAttentionFA} across 4--8 GPUs. We use a learning rate of $1\times10^{-5}$ with cosine annealing, batch size 32, and maximum sequence length 8,192. Notably, training loss proved to be a poor predictor of downstream performance. Models trained on Flash data (lower loss) performed substantially worse than those trained on Pro data on WebArena, consistent with findings in instruction tuning~\citep{Zhou2023LIMALI}. Full hyperparameters are in Appendix~\ref{app:hyperparameters}.

\subsection{Evaluation Setup and Baselines}
\label{sec:experiments}
\label{sec:eval_setup}
\label{sec:baselines}

We evaluate on five benchmarks through BrowserGym~\citep{Chezelles2024TheBE} (details in Appendix~\ref{app:eval_benchmarks}). \textbf{WebArena}~\citep{Zhou2023WebArenaAR} (381 test tasks across six self-hosted web applications) is our \emph{in-domain} benchmark: training data is synthesized on the same environments, but the test tasks are never seen during training and were authored by a separate group of human annotators. \textbf{VisualWebArena}~\citep{Koh2024VisualWebArenaEM} (449 tasks) extends this to vision-grounded tasks. \textbf{WorkArena}~\citep{Drouin2024WorkArenaHC} (L1: 330 episodes) and \textbf{WorkArena++}~\citep{Boisvert2024WorkArenaTC} (L2: 185 tasks) evaluate on ServiceNow, an enterprise platform completely different from any training environment. \textbf{MiniWoB}~\citep{Liu2018ReinforcementLO} (125 task types) tests atomic web skills. All benchmarks except WebArena are \emph{fully out-of-distribution}. All models use a unified token budget of 65,536 tokens for fair comparison. We compare \textbf{proprietary} models (Gemini~3.1~Pro, Gemini~3.1~Flash~Lite, GPT-5~Mini), \textbf{open-weight base} models from the Qwen family~\citep{Bai2025Qwen25VLTR} (Qwen3.5-27B/9B/4B/2B, Qwen3-VL-32B/8B-Thinking), and our fine-tuned \textbf{A3-Qwen3.5-9B}. Additional baselines appear in Appendix~\ref{app:webarena_full}.

\section{Results}
\label{sec:results}

We present cross-benchmark evaluation results (\secref{sec:cross_benchmark}), analyze the role of teacher model quality (\secref{sec:teacher_quality}), and ablate individual pipeline modules (\secref{sec:ablations}).

\subsection{Cross-Benchmark Evaluation}
\label{sec:cross_benchmark}

Our central result is that A3-Qwen3.5-9B, a 9B open-weight model trained with pure SFT on 2{,}322 trajectories, reaches 41.5\% on WebArena, surpassing GPT-4o (31.5\%) and Claude~3.5~Sonnet (36.0\%) under the same GenericAgent/BrowserGym evaluation protocol (Appendix~\ref{app:browsergym_baselines}). This exceeds the previous best open-weight SFT result by a wide margin (Go-Browse~\citep{Gandhi2025GoBrowseTW}, 21.7\%), though that comparison is confounded by differences in agent harness and observation format (Appendix~\ref{app:leaderboard}). As shown in Table~\ref{tab:cross_benchmark}, improvements transfer to all five benchmarks. These out-of-distribution gains confirm that \websynth{} training provides generalizable web interaction capabilities rather than overfitting to WebArena-specific tasks.

\paragraph{Transfer to unseen enterprise interfaces.}
The WorkArena~L1 gain (+18.2pp) is the most striking cross-benchmark result. ServiceNow is an enterprise platform with completely different layouts, form structures, navigation patterns, and design language from any WebArena environment. Yet the student's performance jumps from 33.3\% to 51.5\%, consistent with the hypothesis that the distilled trajectories teach general web interaction primitives (form navigation, table interpretation, field filling) rather than environment-specific shortcuts. WorkArena~L1 tasks (record creation, list filtering, catalog ordering) share structural, though not surface-level, similarity with WebArena tasks like editing wiki pages or managing shopping carts.

The fine-tuned 9B model exactly matches the 3$\times$ larger Qwen3.5-27B on WebArena (41.5\% vs.\ 41.5\%) and is competitive on VisualWebArena (33.9\% vs.\ 37.4\%), though a gap remains on WorkArena++~L2 (9.7\% vs.\ 18.9\%) where longer compositional tasks~\citep{Boisvert2024WorkArenaTC} benefit from larger model capacity.

\begin{figure}[t]
\centering
\includegraphics[width=\linewidth]{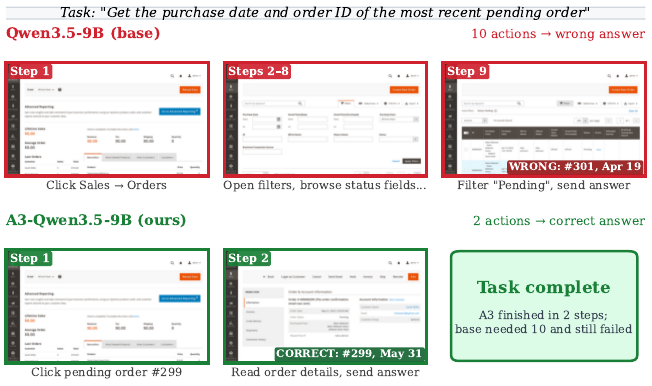}
\caption{Base vs.\ fine-tuned model on a WebArena Shopping Admin task. The base Qwen3.5-9B (top, red) opens filters and browses fields for 10 actions before returning the wrong order. A3-Qwen3.5-9B (bottom, green) clicks the correct pending order from the dashboard and answers in 2 actions.}
\label{fig:qualitative}
\end{figure}

\paragraph{Per-site analysis.}
The per-site breakdown (Table~\ref{tab:webarena} in Appendix~\ref{app:webarena_full}) reveals that the largest absolute gains from fine-tuning come on GitLab (+18.8pp) and Shopping Admin (+18.0pp), which involve complex form-filling and navigation workflows. Distillation benefits also hold at smaller student scales: A3-Qwen3.5-4B gains +11.1pp and A3-Qwen3.5-2B gains +6.1pp on WebArena (Appendix~\ref{app:scaling}), confirming that the distilled data is useful across model capacities with diminishing returns at 2B. Figure~\ref{fig:qualitative} illustrates the behavioral difference. On a Shopping Admin task, the base model wanders through filters for 10 steps and returns the wrong order, while A3-Qwen3.5-9B navigates directly to the correct order in 2 steps. Additional qualitative examples across all five benchmarks are in Appendix~\ref{app:qualitative}.

\subsection{Teacher Model Quality}
\label{sec:teacher_quality}

We compare teacher models by their \websynth{} success rates (Table~\ref{tab:websynth_extended}). Gemini~3~Pro with reduced thinking achieves 69--85\% across environments, substantially above Gemini~3~Flash (17--53\%). Notably, Flash's lower success rate does not mean fewer training examples; failed trajectories tend to be longer (more steps before giving up), so Flash (high thinking) actually produces more observation-action pairs (22,707 vs.\ 16,353 for Pro). Despite this, trajectory quality appears to dominate quantity. A similar pattern appears in InSTA~\citep{Trabucco2025TowardsIT}, where a 1.7B student trained on high-quality data outperformed its 235B teacher.

\paragraph{Lower reasoning budgets produce better trajectories.}
Counterintuitively, reducing the teacher's thinking budget improves \websynth{} success rates across all six environments (Table~\ref{tab:websynth}). While recent work has shown that scaling test-time compute generally improves performance~\citep{Snell2024ScalingLT}, two hypotheses could explain the reversal: (1)~concise reasoning traces may be easier for the student to learn from, presenting cleaner signal with less irrelevant deliberation; (2)~the teacher itself may perform better with a lower reasoning budget, as extended thinking can lead to overthinking and execution errors. Hypothesis (2) is directly supported by our data. The teacher's own success rates on \websynth{} are higher with reduced thinking across all six environments. The detailed comparison of thinking budget configurations and their student outcomes is in Appendix~\ref{app:teacher_comparison}.

\paragraph{Model recency does not predict teaching effectiveness.}
Gemini~3.1~Pro achieves lower \websynth{} success rates than the older Gemini~3~Pro on four of six sites (e.g., 45.4\% vs.\ 78.0\% on Map), indicating that teacher quality \emph{on the specific task distribution} matters more than recency.

\subsection{Ablations}
\label{sec:ablations}

Table~\ref{tab:ablation_main} summarizes ablations of pipeline modules and data scale on WebArena (full details in Appendix~\ref{app:ablations}). The top section ablates modules from the full pipeline (2,322 trajectories). Removing Judge filtering costs 4.5pp despite 40\% more training data, and removing reasoning traces costs 7.9pp. The middle section compares module variants at matched scale (600 tasks each): with-persona data outperforms no-hints data by 2.4pp, confirming that both persona-driven task generation and hint-aided judging contribute to data quality. Performance scales log-linearly with data (Figure~\ref{fig:data_scaling}), with clear diminishing returns. The last 892 trajectories (1,430$\to$2,322) contribute only 1.3pp compared to 5.0pp for the first 430 (285$\to$715), suggesting that substantially more data from the same pipeline would yield only modest additional gains. The framework also enables direct comparison with prior pipelines (e.g., NNetNav, InSTA) by training the same student on data from each; we leave this to future work.

\begin{table}[t]
\centering
\caption{Ablations on WebArena (381 tasks). Top: data scaling and module removal from the full pipeline. Bottom: module variants at matched scale (600 trajectories each).}
\label{tab:ablation_main}
\small
\begin{tabular}{lrcc}
\toprule
\textbf{Ablation} & \textbf{Trajs} & \textbf{SR (\%)} & \textbf{$\Delta$} \\
\midrule
Full pipeline & 2,322 & 41.5 & -- \\
\quad 1,430 trajectories & 1,430 & 40.2 & $-$1.3 \\
\quad 715 trajectories & 715 & 37.0 & $-$4.5 \\
\quad No Judge filtering & 2,999 & 37.0 & $-$4.5 \\
\quad No reasoning traces & 2,322 & 33.6 & $-$7.9 \\
\midrule
With personas, with hints & 600 & 37.8 & -- \\
\quad No hints & 600 & 35.4 & $-$2.4 \\
\bottomrule
\end{tabular}
\end{table}

\section{Discussion}
\label{sec:discussion}

\paragraph{Why six environments suffice.}
Our pipeline uses only six WebArena environments, compared to InSTA's 150,000 websites~\citep{Trabucco2025TowardsIT}. Yet the student transfers across four distinct generalization axes: to an unseen enterprise platform (WorkArena~L1, +18.2pp), to longer-horizon enterprise workflows (WorkArena++~L2, +7.5pp), to visually grounded tasks with image-based instructions (VisualWebArena, +5.4pp), and to simplified HTML micro-tasks (MiniWoB, +5.8pp). We hypothesize that the six environments cover the core web interaction primitives (form filling, table navigation, search, multi-step workflows) that recur across these diverse interfaces. Persona-driven diversity within these environments may be more efficient than scaling to more websites with less task variety per site.

\paragraph{Limitations and scope.}
\label{sec:limitations}
Our ablations validate the Judge, hints, reasoning traces, and data scale, but the persona module lacks a controlled no-persona comparison (Appendix~\ref{app:persona_ablation}); running the full pipeline without personas requires regenerating exploration and task synthesis from scratch, which we leave to future work. The data scaling curve shows diminishing returns at 2,322 trajectories; scaling further within the same six environments would likely require new task generation strategies rather than simply more of the same. The Judge's false positive rate has not been measured against human labels, as building a reliable annotation interface for multi-step web trajectories is a substantial effort in itself; we plan to release the trajectories to enable community validation. All teacher comparisons use the Gemini family because it was the only family offering configurable thinking budgets at the time of our experiments; testing with other families (e.g., Claude, GPT-4) would broaden the generality of our findings. Finally, we use SFT only; combining with RL~\citep{Bai2024DigiRLTI, Putta2024AgentQA} is a natural next step but orthogonal to the data generation focus of this work.

\paragraph{Future directions.}
Several extensions could amplify the gains reported here. Combining our depth-focused approach with broader environment coverage (more websites) would test whether the two strategies are complementary. Self-thinking trace regeneration, which replaces teacher reasoning with student-generated traces while preserving correct actions, could improve the student's reasoning coherence. Iterative self-improvement~\citep{Gulcehre2023ReinforcedS}, where the fine-tuned student generates new trajectories for further training, and RL refinement could compound gains across rounds. The modular structure of \ours{} makes it straightforward to swap in stronger teachers, additional environments, or alternative judge designs as they become available.

\section{Conclusion}
\label{sec:conclusion}

We set out to close the capability gap between frontier LLMs and small open-weight models for web agent tasks. The \ours{} framework, inspired by the roles humans play when annotating web agent benchmarks, provided a structured approach to generating training data with a frontier teacher. The key empirical finding is that data quality matters more than quantity: 2,322 carefully filtered trajectories from a strong teacher suffice to produce a 9B model that surpasses GPT-4o and Claude~3.5~Sonnet on WebArena and transfers to enterprise platforms, visual tasks, and micro-task environments never seen during training. Each pipeline module (Judge filtering, evaluation hints, intact reasoning traces) contributes measurably to this outcome. Perhaps most practically, reducing the teacher's reasoning budget improved both trajectory quality and generation cost, suggesting that capable distillation pipelines need not be expensive. We will release the full trajectory dataset, pipeline code, and model checkpoint to support future work on making web agents broadly accessible.

\section*{Reproducibility Statement}
All experiments use BrowserGym and AgentLab for evaluation. Hyperparameters and training details are specified in \secref{sec:training} and Appendix~\ref{app:hyperparameters}. We will release: (1) the full \websynth{} trajectory dataset (all teacher configurations), (2) the data generation pipeline code, (3) the fine-tuned model checkpoint (A3-Qwen3.5-9B), and (4) the evaluation scripts and configurations.

\section*{Ethics Statement}
Distilling web agent capabilities into small open-weight models lowers the barrier to deploying autonomous web agents, which carries dual-use risk: the same skills that help a user manage their email or shop online could be repurposed for spam, scraping, or other automated interactions that violate website terms of service. We partially mitigate this by training and evaluating exclusively on self-hosted environments (WebArena) and dedicated evaluation instances (WorkArena, MiniWoB), so no production website was affected by our experiments. We plan to release model weights and training data upon acceptance; we encourage the community to develop appropriate safeguards before deploying such agents in production.

\paragraph{Use of AI assistants.} Claude (Anthropic) was used substantially throughout this work: for writing and editing paper content, generating figures and tables, developing data processing and evaluation code, and assisting with experimental analysis. All research ideas, experimental design, and scientific claims were originated and verified by the authors.

\section*{Acknowledgments}
Xing Han L\`{u} acknowledges the support of the Natural Sciences and Engineering Research Council of Canada (NSERC) [funding reference no.\ 579403]. Siva Reddy is supported by a Canada CIFAR AI Chair.

\bibliography{references}
\bibliographystyle{colm2026_conference}

\appendix
\input{appendix}

\end{document}

%% file: math_commands.tex
\usepackage{amsmath,amsfonts,bm}

\def\figref#1{\hyperref[#1]{Figure~\ref*{#1}}}
\def\Figref#1{\hyperref[#1]{Figure~\ref*{#1}}}
\def\twofigref#1#2{Figures \hyperref[#1]{\ref*{#1}} and \hyperref[#2]{\ref*{#2}}}
\def\quadfigref#1#2#3#4{Figures \hyperref[#1]{\ref*{#1}}, \hyperref[#2]{\ref*{#2}}, \hyperref[#3]{\ref*{#3}} and \hyperref[#4]{\ref*{#4}}}
\def\secref#1{\hyperref[#1]{Section~\ref*{#1}}}
\def\Secref#1{\hyperref[#1]{Section~\ref*{#1}}}

\def\eqref#1{equation~\ref{#1}}

\def\1{\bm{1}}

\DeclareMathAlphabet{\mathsfit}{\encodingdefault}{\sfdefault}{m}{sl}
\SetMathAlphabet{\mathsfit}{bold}{\encodingdefault}{\sfdefault}{bx}{n}

%% file: appendix.tex
\section{Extended Related Work}
\label{app:extended_related}

\paragraph{Additional benchmarks and architectures.}
Beyond the benchmarks discussed in the main text, WebShop~\citep{Yao2022WebShopTS} introduced a simulated e-commerce environment, WebCanvas~\citep{Pan2024WebCanvasBW} evaluates agents in live online environments, AssistantBench~\citep{Yoran2024AssistantBenchCW} tests open-web information seeking, TheAgentCompany~\citep{Xu2024TheAgentCompanyBL} evaluates consequential enterprise tasks, and AgentBench~\citep{Liu2023AgentBenchEL} spans multiple agent environments. Web agent architectures have progressed from prompting strategies~\citep{Zheng2024GPT4VisionIA} and modular planning~\citep{Gur2023ARW, Sodhi2023StePSL} to end-to-end vision-language approaches~\citep{Hong2023CogAgentAV, Furuta2023MultimodalWN, He2024WebVoyagerBA, Lai2024AutoWebGLMAL} and tree-search planning~\citep{Yao2023TreeOT, Zhou2023LanguageAT}. BAGEL~\citep{Murty2024BAGELBA} bootstraps agent training via LLM-guided exploration in grounded environments.

\paragraph{Extended knowledge distillation context.}
Synthetic data approaches include evolutionary complexity~\citep{Xu2023WizardLMEL}, direct prompting of aligned models~\citep{Xu2024MagpieAD}, and persona-driven diversity~\citep{Chan2024ScalingSD}. Data quality dominates quantity~\citep{Muennighoff2023ScalingDL}. Zephyr~\citep{Tunstall2023ZephyrDD} showed that distilled alignment (SFT + preference optimization on synthetic data) produces strong chat models. Our Judge filtering is functionally similar to data quality filtering in curation approaches~\citep{Albalak2024ASO, Ding2023EnhancingCL}.

\paragraph{Extended self-improvement context.}
LLM-based evaluation~\citep{Chiang2023CanLL} and RL-based self-improvement through RLHF~\citep{Christiano2017DeepRL, Schulman2017ProximalPO} are complementary to our SFT approach. STaR~\citep{Zelikman2022STaRBR} bootstraps reasoning by training on successful traces, Reflexion~\citep{Shinn2023ReflexionLA} uses verbal self-reflection without weight updates, and Agent Workflow Memory~\citep{Wang2024AgentWM} learns reusable workflows at inference time. Iterative self-improvement~\citep{Madaan2023SelfRefineIR} and combining SFT with RL~\citep{DeepSeek-AI2025DeepSeekR1IR} are promising future directions.

\section{Extended Results and Analysis}
\label{app:extended_results}

\subsection{\websynth{} Teacher Model Results}
\label{app:websynth_teacher}
\label{app:websynth_extended}
\label{app:teacher_comparison}

Table~\ref{tab:websynth} shows the per-site success rates and training example counts for all teacher model configurations on \websynth{} tasks. Each teacher runs 500 tasks per site. Success rates are determined by the Judge module. The ``Examples'' column shows the total number of observation-action training pairs extracted from successful trajectories across all six sites.

\begin{table}[h]
\centering
\caption{Pro (reduced thinking) achieves the highest success rates across all environments, though Flash produces more total training examples (22,707 vs.\ 16,353) from lower-quality trajectories. Extended teacher model comparison on \websynth{}: success rate (\%) by site and total training examples produced.}
\label{tab:websynth_extended}
\label{tab:websynth}
\small
\setlength{\tabcolsep}{3pt}
\begin{tabular}{lccccccc}
\toprule
\textbf{Teacher} & \textbf{Reddit} & \textbf{GitLab} & \textbf{Shopping} & \textbf{Shop Adm.} & \textbf{Wikipedia} & \textbf{Map} & \textbf{Examples} \\
\midrule
\multicolumn{8}{l}{\textit{Gemini 3 Pro}} \\
\textbf{Pro (reduced thinking)} & \textbf{69.0} & \textbf{70.6} & \textbf{80.8} & \textbf{80.8} & \textbf{85.4} & \textbf{78.0} & \textbf{16,353} \\
Pro & 66.0 & 64.2 & 68.8 & 78.2 & 78.8 & 74.4 & 15,351 \\
\midrule
\multicolumn{8}{l}{\textit{Gemini 3.1 Pro}} \\
3.1 Pro (reduced thinking) & 58.4 & 67.6 & 61.4 & 63.8 & 51.0 & 45.4 & 14,859 \\
\midrule
\multicolumn{8}{l}{\textit{Gemini 3 Flash}} \\
Flash (high thinking) & 45.8 & 33.0 & 39.4 & 39.4 & 35.6 & 17.0 & 22,707 \\
Flash (reduced thinking) & 45.8 & 35.6 & 40.2 & 38.4 & 33.6 & 16.4 & 10,943 \\
\bottomrule
\end{tabular}
\end{table}

Several key observations emerge from this comparison:

\begin{enumerate}[nosep,leftmargin=1.5em]
\item \textbf{Pro (reduced thinking) dominates across all sites.} The reduced thinking budget configuration produces concise yet effective reasoning traces, achieving the highest success rates on every environment (69--85\%).
\item \textbf{Gemini 3.1 Pro underperforms Gemini 3 Pro.} Despite being a newer model, Gemini 3.1 Pro (reduced thinking) achieves lower success rates on five of six sites, with a particularly large drop on Map (45.4\% vs.\ 78.0\%). This suggests that the newer model's capabilities may not transfer uniformly to self-hosted web environments.
\item \textbf{Flash models show a ceiling around 50\%.} All Flash configurations fall below 54\% on every site, with Map being especially challenging ($\leq$24.1\%).
\item \textbf{Training examples do not directly track success rate.} Flash produces the most training examples (22,707) due to longer trajectories, but these come from lower-quality task completions. Pro (reduced thinking) produces 16,353 examples with higher average quality.
\end{enumerate}

\begin{figure}[h]
\centering
\includegraphics[width=\linewidth]{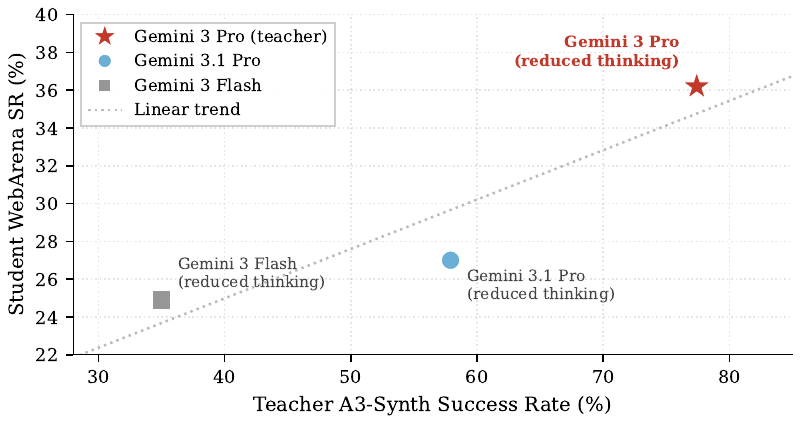}
\caption{Teacher quality on \websynth{} (x-axis) vs.\ student performance on WebArena (y-axis) across three teacher configurations. The student is Qwen3-VL-8B-Thinking, an earlier-generation model that was trained on all three teacher variants before Qwen3.5-9B became available. Based on these results, we selected Gemini~3~Pro (reduced thinking) as the teacher for the primary A3-Qwen3.5-9B model (41.5\%).}
\label{fig:teacher_quality}
\end{figure}

\paragraph{Teacher selection.} To validate that these \websynth{} quality differences translate to downstream student performance, we trained Qwen3-VL-8B-Thinking on data from three teacher configurations (Gemini~3~Pro, Gemini~3.1~Pro, and Gemini~3~Flash, all with reduced thinking) and evaluated each on WebArena (\figref{fig:teacher_quality}). The Pro-trained student (36.2\%) substantially outperformed the 3.1~Pro-trained (27.0\%) and Flash-trained (24.9\%) variants, confirming that teacher quality on \websynth{} predicts student performance. Based on these results, we did not pursue Flash or 3.1~Pro further and focused all subsequent experiments on Gemini~3~Pro (reduced thinking) as the teacher.

\subsection{Per-Benchmark Detailed Results}
\label{app:per_benchmark}
\label{app:cross_benchmark_fig}

\begin{figure}[h]
\centering
\includegraphics[width=\linewidth]{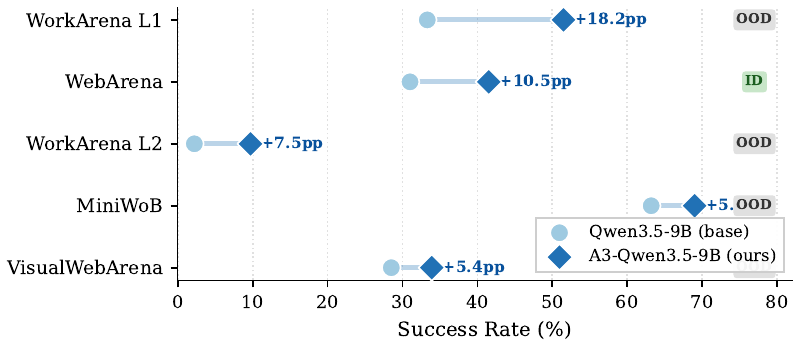}
\caption{Cross-benchmark success rates for Qwen3.5-9B before and after fine-tuning on \websynth{} Pro (reduced thinking) trajectories. The largest gain is on WorkArena~L1 (+18.2pp), an enterprise interface never seen during training. All benchmarks except WebArena are fully out-of-distribution.}
\label{fig:cross_benchmark}
\end{figure}

All results in this section use the official BrowserGym test/train splits~\citep{Chezelles2024TheBE}. For WebArena (381 test / 431 train), VisualWebArena (449 test / 461 train), and WorkArena~L2 (185 test / 156 train), we report on the \emph{test split only} to avoid training data contamination. For WorkArena~L1 and MiniWoB, we use the full task set (matching the BrowserGym evaluation protocol). Figure~\ref{fig:cross_benchmark} visualizes the improvements from Table~\ref{tab:cross_benchmark} across all five benchmarks.

\paragraph{WebArena.}
\label{app:webarena_full}

Table~\ref{tab:webarena} shows the per-site breakdown across all evaluated models. We evaluate on 381 tasks across six self-hosted web applications.

\begin{table}[h]
\centering
\caption{Fine-tuned Qwen3.5-9B (41.5\%) matches the 3$\times$ larger Qwen3.5-27B on WebArena, with the largest gains on GitLab (+18.8pp) and Shopping Admin (+18.0pp). Per-site success rates (\%); \textbf{bold} = best open-weight $\leq$9B.}
\label{tab:webarena}
\label{tab:webarena_full}
\small
\setlength{\tabcolsep}{3.5pt}
\begin{tabular}{lcccccc}
\toprule
\textbf{Model} & \textbf{Reddit} & \textbf{GitLab} & \textbf{Shopping} & \textbf{Shop Admin} & \textbf{Map} & \textbf{All} \\
\midrule
\multicolumn{7}{l}{\textit{Proprietary frontier models}} \\
Gemini 3.1 Pro & 79.2 & 57.4 & 43.3 & 55.1 & 42.2 & 53.8 \\
Gemini 3 Pro & 75.0 & 56.4 & 34.4 & 53.8 & 45.3 & 51.2 \\
Gemini 3.1 Flash Lite & 68.8 & 48.5 & 32.2 & 41.0 & 28.1 & 42.3 \\
\midrule
\multicolumn{7}{l}{\textit{Open-weight models (base)}} \\
Qwen3.5-27B & 56.2 & 44.6 & 35.6 & 46.2 & 28.1 & 41.5 \\
Qwen3.5-9B & 54.2 & 30.7 & 30.0 & 26.9 & 20.3 & 31.0 \\
Qwen3.5-4B & 33.3 & 23.8 & 24.4 & 28.2 & 12.5 & 24.1 \\
Qwen3.5-2B & 4.2 & 2.0 & 7.8 & 1.3 & 0.0 & 3.1 \\
\midrule
\multicolumn{7}{l}{\textit{A3 fine-tuned (ours)}} \\
\textbf{A3-Qwen3.5-9B} & \textbf{56.2} {\scriptsize\textcolor{teal}{+2.0}} & \textbf{49.5} {\scriptsize\textcolor{teal}{+18.8}} & \textbf{33.3} {\scriptsize\textcolor{teal}{+3.3}} & \textbf{44.9} {\scriptsize\textcolor{teal}{+18.0}} & \textbf{25.0} {\scriptsize\textcolor{teal}{+4.7}} & \textbf{41.5} {\scriptsize\textcolor{teal}{+10.5}} \\
A3-Qwen3.5-4B & 50.0 {\scriptsize\textcolor{teal}{+16.7}} & 39.6 {\scriptsize\textcolor{teal}{+15.8}} & 32.2 {\scriptsize\textcolor{teal}{+7.8}} & 30.0 {\scriptsize\textcolor{teal}{+1.8}} & 26.4 {\scriptsize\textcolor{teal}{+13.9}} & 35.2 {\scriptsize\textcolor{teal}{+11.1}} \\
A3-Qwen3.5-2B & 6.2 {\scriptsize\textcolor{teal}{+2.0}} & 6.9 {\scriptsize\textcolor{teal}{+4.9}} & 14.4 {\scriptsize\textcolor{teal}{+6.6}} & 7.5 {\scriptsize\textcolor{teal}{+6.2}} & 11.3 {\scriptsize\textcolor{teal}{+11.3}} & 9.2 {\scriptsize\textcolor{teal}{+6.1}} \\
\bottomrule
\end{tabular}
\end{table}

\paragraph{VisualWebArena.}
\label{app:vwa_results}

Table~\ref{tab:vwa_full} reports results on VisualWebArena, which requires visual understanding of web content across classifieds, shopping, and Reddit sites. For Qwen3.5-9B and A3-Qwen3.5-9B, we report results on the full benchmark (910 tasks, merging the test and train splits); other models are evaluated on the test split only (449 tasks). Our fine-tuned A3-Qwen3.5-9B achieves 33.7\% on the full 910-task benchmark, a +7.5pp improvement over the base model (26.2\%). The model receives both accessibility tree observations and screenshots during evaluation, leveraging its multimodal capabilities for visually grounded tasks.

\begin{table}[h]
\centering
\caption{VisualWebArena results. Qwen3.5-9B and A3-Qwen3.5-9B are evaluated on the full 910 tasks (test + train); other models use the test split only (449 tasks, marked $^\dagger$). Success rate (\%) by site. Cross-site tasks are grouped by starting site.}
\label{tab:vwa_full}
\small
\setlength{\tabcolsep}{8pt}
\begin{tabular}{lcccc}
\toprule
\textbf{Model} & \textbf{Classifieds} & \textbf{Shopping} & \textbf{Reddit} & \textbf{All} \\
\midrule
\multicolumn{5}{l}{\textit{Proprietary frontier models}} \\
Gemini 3.1 Pro & 47.4 & 51.9 & 39.0 & 47.9 \\
Gemini 3 Pro & 51.7 & 49.8 & 44.0 & 49.0 \\
Gemini 3 Flash & 37.9 & 40.8 & 35.0 & 38.8 \\
Gemini 3.1 Flash Lite & 31.9 & 36.9 & 34.0 & 35.0 \\
\midrule
\multicolumn{5}{l}{\textit{Open-weight models (base)}} \\
Qwen3.5-27B & 37.9 & 39.9 & 31.0 & 37.4 \\
Qwen3.5-9B$^\dagger$ & 22.4 & 32.6 & 20.0 & 26.2 \\
Qwen3-VL-8B-Thinking & 25.0 & 30.0 & 16.0 & 25.6 \\
Qwen3.5-4B & 29.8 & 27.6 & 16.5 & 24.7 \\
Qwen3.5-2B & 5.3 & 6.1 & 5.1 & 5.3 \\
\midrule
\multicolumn{5}{l}{\textit{A3 fine-tuned (ours)}} \\
\textbf{A3-Qwen3.5-9B}$^\dagger$ & \textbf{35.8} {\scriptsize\textcolor{teal}{+13.4}} & \textbf{35.2} {\scriptsize\textcolor{teal}{+2.6}} & \textbf{34.1} {\scriptsize\textcolor{teal}{+14.1}} & \textbf{33.7} {\scriptsize\textcolor{teal}{+7.5}} \\
\textbf{A3-Qwen3.5-4B} & 28.1 {\scriptsize\textcolor{teal}{-1.7}} & 34.6 {\scriptsize\textcolor{teal}{+7.0}} & 26.6 {\scriptsize\textcolor{teal}{+10.1}} & \textbf{30.1} {\scriptsize\textcolor{teal}{+5.4}} \\
\textbf{A3-Qwen3.5-2B} & 7.9 {\scriptsize\textcolor{teal}{+2.6}} & 8.3 {\scriptsize\textcolor{teal}{+2.2}} & 6.3 {\scriptsize\textcolor{teal}{+1.2}} & \textbf{7.6} {\scriptsize\textcolor{teal}{+2.2}} \\
\bottomrule
\end{tabular}
\end{table}

\paragraph{WorkArena L1.}
\label{app:workarena_l1}

Table~\ref{tab:workarena_l1_full} presents results on WorkArena L1, which consists of 330 tasks on ServiceNow enterprise software. Tasks test basic operations such as creating records, filtering lists, sorting columns, ordering items, reading charts, and basic navigation. Our fine-tuned 9B model achieves 51.5\%, a +18.2pp absolute improvement over the base Qwen3.5-9B (33.3\%), narrowing the gap to the larger Qwen3.5-27B (57.0\%).

\begin{table}[h]
\centering
\caption{A3 fine-tuning produces the largest gain on WorkArena L1 (+18.2pp), with the biggest per-category jump on Order tasks (+38.9pp). Success rate (\%) by task category; Crt = Create (50), Flt = Filter (60), Srt = Sort (60), Ord = Order (90), Cht = Chart (40), Nav = Navigation (30).}
\label{tab:workarena_l1_full}
\small
\setlength{\tabcolsep}{3pt}
\begin{tabular}{lcccccc c}
\toprule
\textbf{Model} & \textbf{Create} & \textbf{Filter} & \textbf{Sort} & \textbf{Order} & \textbf{Chart} & \textbf{Navigate} & \textbf{All} \\
\midrule
\multicolumn{8}{l}{\textit{Proprietary frontier models}} \\
Gemini 3.1 Pro & 70.0 & 33.3 & 86.7 & 98.9 & 90.0 & 100 & 79.4 \\
Gemini 3 Pro & 72.0 & 35.0 & 88.3 & 100 & 82.5 & 100 & 79.7 \\
Gemini 3.1 Flash Lite & 52.0 & 31.7 & 10.0 & 94.4 & 70.0 & 96.7 & 58.5 \\
\midrule
\multicolumn{8}{l}{\textit{Open-weight models (base)}} \\
Qwen3.5-27B & 60.0 & 13.3 & 11.7 & 95.6 & 70.0 & 96.7 & 57.0 \\
Qwen3-VL-8B-Thinking & 40.0 & 1.7 & 10.0 & 83.3 & 57.5 & 60.0 & 43.3 \\
Qwen3.5-4B & 38.0 & 1.7 & 11.7 & 45.6 & 60.0 & 63.3 & 33.6 \\
Qwen3.5-9B & 32.0 & 1.7 & 13.3 & 57.8 & 42.5 & 53.3 & 33.3 \\
Qwen3.5-2B & 0.0 & 0.0 & 11.7 & 1.1 & 12.5 & 3.3 & 4.2 \\
\midrule
\multicolumn{8}{l}{\textit{A3 fine-tuned (ours)}} \\
\textbf{A3-Qwen3.5-9B} & \textbf{46.0} {\scriptsize\textcolor{teal}{+14.0}} & 1.7 & \textbf{20.0} {\scriptsize\textcolor{teal}{+6.7}} & \textbf{96.7} {\scriptsize\textcolor{teal}{+38.9}} & \textbf{65.0} {\scriptsize\textcolor{teal}{+22.5}} & \textbf{70.0} {\scriptsize\textcolor{teal}{+16.7}} & \textbf{51.5} {\scriptsize\textcolor{teal}{+18.2}} \\
\textbf{A3-Qwen3.5-4B} & 40.0 {\scriptsize\textcolor{teal}{+2.0}} & 0.0 {\scriptsize\textcolor{teal}{-1.7}} & 11.7 & 92.2 {\scriptsize\textcolor{teal}{+46.6}} & 57.5 {\scriptsize\textcolor{teal}{-2.5}} & 50.0 {\scriptsize\textcolor{teal}{-13.3}} & \textbf{44.8} {\scriptsize\textcolor{teal}{+11.2}} \\
\textbf{A3-Qwen3.5-2B} & 0.0 & 0.0 & 0.0 {\scriptsize\textcolor{teal}{-11.7}} & 4.4 {\scriptsize\textcolor{teal}{+3.3}} & 32.5 {\scriptsize\textcolor{teal}{+20.0}} & 16.7 {\scriptsize\textcolor{teal}{+13.4}} & \textbf{6.7} {\scriptsize\textcolor{teal}{+2.5}} \\
\bottomrule
\end{tabular}
\end{table}

\paragraph{WorkArena++ L2.}
\label{app:workarena_l2}

Table~\ref{tab:workarena_l2_full} presents results on WorkArena++~\citep{Boisvert2024WorkArenaTC} L2, which consists of 185 tasks requiring multi-step, compositional reasoning on ServiceNow. Tasks involve dashboard interpretation, filtering with problem identification, expense management, work assignment, multi-channel workflows, and navigation. These tasks are substantially harder than L1, requiring longer interaction horizons (up to 50 steps) and compositional understanding.

Our fine-tuned A3-Qwen3.5-9B achieves 9.7\%, a +7.5pp improvement over the base model (2.2\%), representing a 4.4$\times$ relative improvement. The large gap to Gemini 3.1 Pro (40.0\%) indicates that L2-level compositional enterprise tasks~\citep{Boisvert2024WorkArenaTC} remain challenging for small models and may require enterprise-specific training data.

\begin{table}[h]
\centering
\caption{A3 fine-tuning yields a 4.4$\times$ relative improvement on WorkArena++ L2 (2.2\% $\rightarrow$ 9.7\%), with gains concentrated on Filter tasks. Dashboard tasks (n=57) are 0\% for all models. Success rate (\%) by category; Dsh = Dashboard (57), Flt = Filter (42), Nav = Navigate (13), Inf = Infeasible (26), Oth = Other (47).}
\label{tab:workarena_l2_full}
\small
\begin{tabular}{lccccc c}
\toprule
\textbf{Model} & \textbf{Dashboard} & \textbf{Filter} & \textbf{Navigate} & \textbf{Infeasible} & \textbf{Other} & \textbf{All} \\
\midrule
\multicolumn{7}{l}{\textit{Proprietary frontier models}} \\
Gemini 3.1 Pro & 0.0 & 76.2 & 53.8 & 53.8 & 29.8 & 40.0 \\
Gemini 3 Pro & 0.0 & 81.0 & 53.8 & 50.0 & 40.0 & 41.6 \\
Gemini 3.1 Flash Lite & 0.0 & 69.0 & 30.8 & 11.5 & 6.4 & 21.1 \\
\midrule
\multicolumn{7}{l}{\textit{Open-weight models (base)}} \\
Qwen3.5-27B & 0.0 & 31.0 & 15.4 & 50.0 & 10.6 & 18.9 \\
Qwen3.5-9B & 0.0 & 2.4 & 15.4 & 3.8 & 0.0 & 2.2 \\
Qwen3-VL-8B-Thinking & 0.0 & 0.0 & 0.0 & 7.7 & 0.0 & 1.1 \\
Qwen3.5-4B & 0.0 & 4.8 & 7.7 & 0.0 & 0.0 & 1.6 \\
Qwen3.5-2B & 0.0 & 0.0 & 0.0 & 0.0 & 0.0 & 0.0 \\
\midrule
\multicolumn{7}{l}{\textit{A3 fine-tuned (ours)}} \\
\textbf{A3-Qwen3.5-9B} & 0.0 & \textbf{33.3} {\scriptsize\textcolor{teal}{+30.9}} & \textbf{23.1} {\scriptsize\textcolor{teal}{+7.7}} & 0.0 & 4.3 {\scriptsize\textcolor{teal}{+4.3}} & \textbf{9.7} {\scriptsize\textcolor{teal}{+7.5}} \\
\textbf{A3-Qwen3.5-4B} & 0.0 & 7.1 {\scriptsize\textcolor{teal}{+2.3}} & 30.8 {\scriptsize\textcolor{teal}{+23.1}} & 0.0 & 0.0 & \textbf{3.8} {\scriptsize\textcolor{teal}{+2.2}} \\
A3-Qwen3.5-2B & 0.0 & 0.0 & 0.0 & 0.0 & 0.0 & 0.0 \\
\bottomrule
\end{tabular}
\end{table}

\paragraph{MiniWoB.}
\label{app:miniwob}

Table~\ref{tab:miniwob_full} presents results on MiniWoB (625 tasks), a benchmark of simplified web interaction tasks testing atomic skills such as clicking, typing, selecting, and basic form-filling. MiniWoB is fully out-of-distribution from our WebArena-based training data; improvements here reflect the acquisition of generalizable low-level web interaction skills.

\begin{table}[h]
\centering
\caption{A3 fine-tuning improves atomic web skills on MiniWoB (+5.8pp), with the largest gains on Type (+11.4pp) and Workflow (+10.6pp) tasks. Success rate (\%) by category; Clk = Click (160), Typ = Type (123), Drg = Drag (55), Wkf = Workflow (85), Wdg = Widget (59), Cmp = Comprehend (52), Rsn = Reason (91).}
\label{tab:miniwob_full}
\scriptsize
\setlength{\tabcolsep}{2.5pt}
\begin{tabular}{lccccccc c}
\toprule
\textbf{Model} & \textbf{Click} & \textbf{Type} & \textbf{Drag} & \textbf{Wkfl} & \textbf{Widget} & \textbf{Compr.} & \textbf{Reason} & \textbf{All} \\
\midrule
\multicolumn{9}{l}{\textit{Proprietary frontier models}} \\
Gemini 3.1 Pro & 87.5 & 84.6 & 16.4 & 98.8 & 83.1 & 71.2 & 64.8 & 77.1 \\
Gemini 3 Pro & 83.8 & 83.2 & 16.4 & 91.8 & 81.7 & 70.9 & 63.5 & 74.7 \\
Gemini 3.1 Flash Lite & 85.0 & 79.7 & 7.3 & 95.3 & 79.7 & 73.1 & 64.8 & 74.1 \\
\midrule
\multicolumn{9}{l}{\textit{Open-weight models (base)}} \\
Qwen3.5-9B & 80.0 & 61.0 & 1.8 & 77.6 & 55.9 & 61.5 & 65.9 & 63.2 \\
Qwen3.5-4B & 78.8 & 52.8 & 5.5 & 74.1 & 61.0 & 57.7 & 64.8 & 61.1 \\
Qwen3.5-2B & 18.8 & 4.0 & 0.0 & 15.7 & 0.0 & 5.7 & 11.8 & 11.8 \\
\midrule
\multicolumn{9}{l}{\textit{A3 fine-tuned (ours)}} \\
\textbf{A3-Qwen3.5-9B} & \textbf{81.2} {\scriptsize\textcolor{teal}{+1.2}} & \textbf{72.4} {\scriptsize\textcolor{teal}{+11.4}} & \textbf{5.5} {\scriptsize\textcolor{teal}{+3.7}} & \textbf{88.2} {\scriptsize\textcolor{teal}{+10.6}} & \textbf{62.7} {\scriptsize\textcolor{teal}{+6.8}} & \textbf{63.5} {\scriptsize\textcolor{teal}{+2.0}} & \textbf{70.3} {\scriptsize\textcolor{teal}{+4.4}} & \textbf{69.0} {\scriptsize\textcolor{teal}{+5.8}} \\
\textbf{A3-Qwen3.5-4B} & 82.5 {\scriptsize\textcolor{teal}{+3.7}} & 68.8 {\scriptsize\textcolor{teal}{+16.0}} & 1.8 {\scriptsize\textcolor{teal}{-3.7}} & 88.2 {\scriptsize\textcolor{teal}{+14.1}} & 63.3 {\scriptsize\textcolor{teal}{+3.3}} & 63.6 {\scriptsize\textcolor{teal}{+3.6}} & 60.0 {\scriptsize\textcolor{teal}{-4.7}} & \textbf{66.9} {\scriptsize\textcolor{teal}{+5.8}} \\
\textbf{A3-Qwen3.5-2B} & 44.1 {\scriptsize\textcolor{teal}{+25.3}} & 20.0 {\scriptsize\textcolor{teal}{+16.0}} & 0.0 & 30.0 {\scriptsize\textcolor{teal}{+14.3}} & 60.0 {\scriptsize\textcolor{teal}{+60.0}} & 31.4 {\scriptsize\textcolor{teal}{+25.7}} & 40.0 {\scriptsize\textcolor{teal}{+28.2}} & \textbf{38.6} {\scriptsize\textcolor{teal}{+26.8}} \\
\bottomrule
\end{tabular}
\end{table}

\subsection{Qualitative Comparison: Base vs.\ Fine-Tuned}
\label{app:qualitative}

We present qualitative examples from each of the five evaluation benchmarks, comparing the base Qwen3.5-9B with the fine-tuned A3-Qwen3.5-9B. Each figure shows key steps from both models on the same task, where the base model fails and the A3 model succeeds.

\paragraph{WebArena.}
Figure~\ref{fig:qualitative} shows a Shopping Admin task requiring the most recent pending order. The base model wanders through menus for 10 actions, opens filters, and returns the wrong order. The A3 model navigates directly to the correct order in 2 actions, having learned efficient navigation patterns from the \websynth{} training data.

\begin{figure}[h]
\centering
\includegraphics[width=\linewidth]{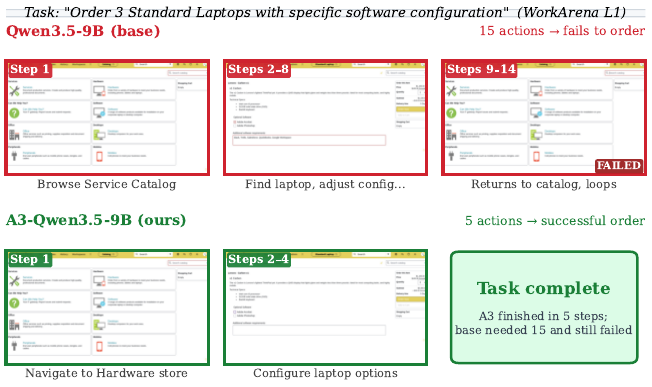}
\caption{Base vs.\ A3 model on WorkArena L1 (order Standard Laptop). The base model (top, red) finds the product page but fails to configure and submit the order within 15 actions. The A3 model (bottom, green) efficiently fills in all required fields and places the order in 5 actions.}
\label{fig:qualitative_workarena_l1}
\end{figure}

\paragraph{WorkArena L1.}
Figure~\ref{fig:qualitative_workarena_l1} shows a ServiceNow hardware ordering task. The base model browses the catalog and finds the laptop page but loops through configuration options for 15 actions without placing the order. The A3 model navigates to the same page, correctly fills in the required software configuration and quantity, and completes the order in 5 actions.

\begin{figure}[h]
\centering
\includegraphics[width=\linewidth]{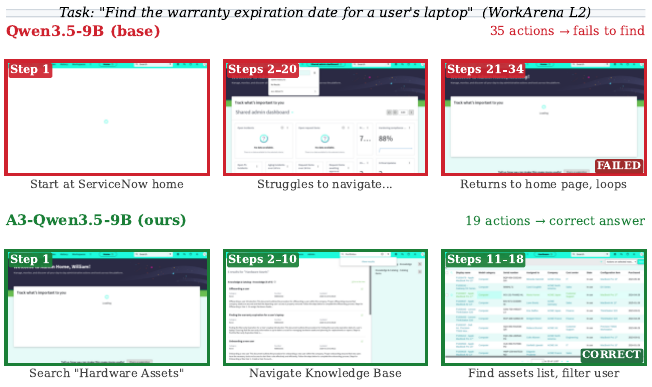}
\caption{Base vs.\ A3 model on WorkArena L2 (find warranty expiration date). The base model (top, red) fails to navigate to the hardware assets module within 35 actions. The A3 model (bottom, green) navigates through the knowledge base, finds the assets list, filters by user, and reports the correct warranty date.}
\label{fig:qualitative_workarena_l2}
\end{figure}

\paragraph{WorkArena L2.}
Figure~\ref{fig:qualitative_workarena_l2} shows a multi-step ServiceNow task requiring navigation to an asset management module, applying filters, and extracting a warranty date. The base model struggles with ServiceNow's navigation for 35 actions, repeatedly returning to the home page without reaching the hardware assets list. The A3 model systematically searches for the correct module, applies the user filter, and reports the correct warranty date in 19 actions.

\begin{figure}[h]
\centering
\includegraphics[width=\linewidth]{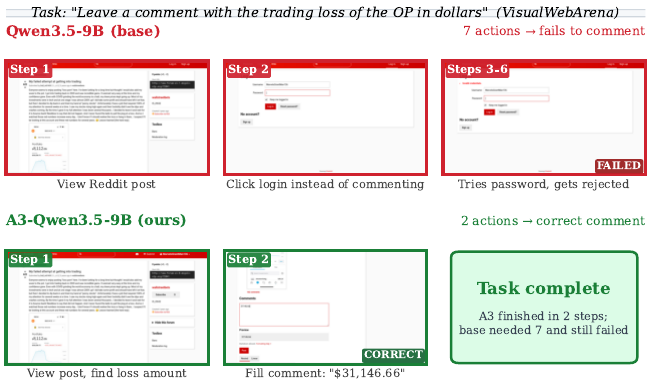}
\caption{Base vs.\ A3 model on VisualWebArena task 378 (Reddit comment). The base model (top, red) clicks login and fails authentication for 7 actions. The A3 model (bottom, green) directly types the trading loss amount into the comment box and submits in 2 actions.}
\label{fig:qualitative_vwa}
\end{figure}

\paragraph{VisualWebArena.}
Figure~\ref{fig:qualitative_vwa} shows a Reddit-like forum task requiring the model to read a post about trading losses and leave a comment with the dollar amount. The base model unnecessarily clicks the login button and gets stuck on the login page (entering incorrect credentials for 5 actions), never returning to the post. The A3 model reads the post, extracts the loss amount, and directly fills in the comment field in 2 actions.

\begin{figure}[h]
\centering
\includegraphics[width=\linewidth]{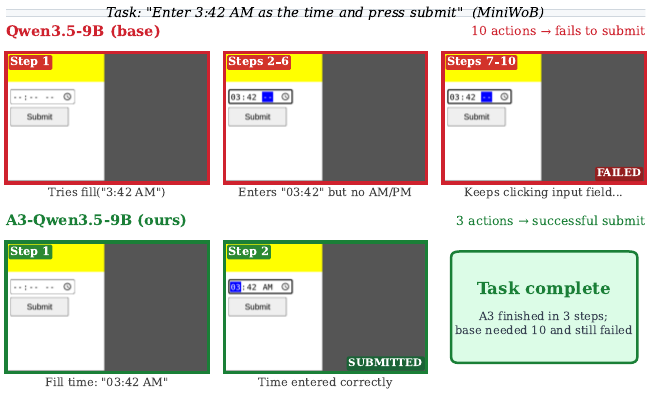}
\caption{Base vs.\ A3 model on MiniWoB (enter time). The base model (top, red) repeatedly clicks the input field for 10 actions without entering the time correctly. The A3 model (bottom, green) fills in ``03:42 AM'' in the correct format and submits in 3 actions.}
\label{fig:qualitative_miniwob}
\end{figure}

\paragraph{MiniWoB.}
Figure~\ref{fig:qualitative_miniwob} shows a time entry task. The base model attempts to fill the time input but uses an incorrect format and never selects the AM/PM field, spending 10 actions clicking the same input field repeatedly. The A3 model enters the time in the correct format (``03:42 AM'') and clicks Submit in 3 actions, demonstrating learned knowledge of HTML form input conventions.

\subsection{Scale Interactions}
\label{app:scaling}

Distillation benefits hold across three student scales: A3-Qwen3.5-9B gains +10.5pp, A3-Qwen3.5-4B gains +11.1pp, and A3-Qwen3.5-2B gains +6.1pp on WebArena (per-site breakdowns in Table~\ref{tab:webarena}). The 4B model shows comparable gains to 9B, while the 2B model shows diminishing returns, likely due to limited model capacity. At 9B, the fine-tuned model (41.5\%) matches the 27B base model (41.5\%), bridging a 3$\times$ parameter gap. Whether the already-strong Qwen3.5-27B would still benefit from distillation remains an open question.

\subsection{Comparison with BrowserGym Ecosystem Results}
\label{app:browsergym_baselines}

A key advantage of building on BrowserGym~\citep{Chezelles2024TheBE} and AgentLab is that our evaluation uses the \emph{same agent harness, observation format, and evaluation protocol} as the broader BrowserGym ecosystem. This enables meaningful comparison with results from the AgentLab TMLR evaluation traces, publicly available on HuggingFace (\texttt{agentlabtraces/agentlabtraces}), which cover six models: Claude~3.5~Sonnet, GPT-4o, GPT-4o-mini, o1-mini, Llama~3.1~405B, and Llama~3.1~70B.

\paragraph{Why this comparison matters.} Most cross-paper comparisons of web agents are confounded by differences in agent prompts, action spaces, observation formats, and evaluation infrastructure. Because both our evaluation and the BrowserGym TMLR traces use the same standardized GenericAgent pipeline with accessibility-tree observations and the same programmatic reward functions, differences in success rates primarily reflect differences in model capability rather than implementation artifacts. This makes BrowserGym the closest available reference for fair comparison.

\paragraph{Evaluation overlap.} For benchmarks with official test/train splits (WebArena, VisualWebArena, WorkArena~L2), we compare on the \emph{test split only}, matching our evaluation protocol. For our primary models (Qwen3.5-9B base and A3-Qwen3.5-9B), we additionally evaluate on the train splits for WebArena and VisualWebArena, reporting the merged full-benchmark results in Tables~\ref{tab:webarena_full} and~\ref{tab:vwa_full}. For WorkArena~L1 and MiniWoB, both evaluations use the full task set.

One exception is WorkArena~L2: the TMLR traces were collected with an older WorkArena version that used different task parametrizations (e.g., ``developer-laptop'' vs.\ ``development-laptop-p-c''). Only 147 of our 185 test tasks overlap with the TMLR task set. We report L2 numbers on this 147-task subset and mark them accordingly.

\begin{table}[h]
\centering
\caption{Comparison with BrowserGym ecosystem results using the same agent harness and evaluation protocol. A3-Qwen3.5-9B exceeds GPT-4o and Claude~3.5~Sonnet on WebArena, approaches frontier models on WorkArena~L1, and matches Claude~3.5~Sonnet on MiniWoB. Success rate (\%). $\dagger$ = 147/185 task overlap due to WorkArena version mismatch.}
\label{tab:browsergym_baselines}
\small
\setlength{\tabcolsep}{4pt}
\begin{tabular}{llccccc}
\toprule
\textbf{Model} & \textbf{Params} & \textbf{WA} & \textbf{VWA} & \textbf{L1} & \textbf{L2} & \textbf{WoB} \\
\midrule
\multicolumn{7}{l}{\textit{BrowserGym ecosystem (same harness)}} \\
Claude 3.5 Sonnet & -- & 36.0 & 22.0 & 56.4 & 38.8$^\dagger$ & 69.8 \\
o1-mini & -- & 29.9 & -- & 56.7 & 14.3$^\dagger$ & 67.8 \\
GPT-4o & -- & 31.5 & 26.3 & 45.5 & 7.5$^\dagger$ & 63.8 \\
Llama 3.1 405B & 405B & 22.6 & -- & 43.3 & 8.9$^\dagger$ & 64.6 \\
Llama 3.1 70B & 70B & 17.1 & -- & 27.9 & 3.4$^\dagger$ & 57.6 \\
GPT-4o-mini & -- & 13.6 & 18.0 & 27.0 & 2.0$^\dagger$ & 56.6 \\
\midrule
\multicolumn{7}{l}{\textit{This work (same harness)}} \\
Gemini 3.1 Pro & -- & 53.8 & 47.9 & 79.4 & 40.0 & 77.1 \\
Qwen3.5-27B & 27B & 41.5 & 37.4 & 57.0 & 18.9 & 70.9 \\
\textbf{A3-Qwen3.5-9B} & \textbf{9B} & \textbf{41.5} & \textbf{33.9} & \textbf{51.5} & \textbf{9.7} & \textbf{69.0} \\
Qwen3.5-9B (base) & 9B & 31.0 & 28.5 & 33.3 & 2.2 & 63.2 \\
\bottomrule
\end{tabular}
\end{table}

\paragraph{Key findings.} Table~\ref{tab:browsergym_baselines} reveals several patterns:

\begin{enumerate}[nosep,leftmargin=1.5em]
\item \textbf{A3-Qwen3.5-9B exceeds GPT-4o and Claude~3.5~Sonnet on WebArena.} Our fine-tuned 9B model (41.5\%) surpasses GPT-4o (31.5\%) by 10.0pp and Claude~3.5~Sonnet (36.0\%) by 5.5pp, despite being a small open-weight model evaluated under the same protocol.

\item \textbf{WorkArena~L1 approaches frontier models.} A3-Qwen3.5-9B (51.5\%) falls between GPT-4o (45.5\%) and Claude~3.5~Sonnet (56.4\%) on an enterprise interface never seen during training.

\item \textbf{MiniWoB matches Claude~3.5~Sonnet.} Our fine-tuned model (69.0\%) closely matches Claude~3.5~Sonnet (69.8\%) on atomic web interaction tasks.

\item \textbf{Fine-tuning closes the gap to much larger models.} The base Qwen3.5-9B performs comparably to GPT-4o-mini and Llama~3.1~70B across benchmarks. After fine-tuning on \websynth{} data, it surpasses models 7--45$\times$ larger (Llama~3.1~70B, GPT-4o, Llama~3.1~405B) on WebArena.
\end{enumerate}

\paragraph{Remaining caveats.} While the shared harness eliminates most confounds, two differences remain: (1) the BrowserGym traces were collected 6--12 months earlier with an older BrowserGym version, and minor changes in observation processing or action parsing may affect results; (2) our evaluation uses newer model releases (Gemini~3.1~Pro, Qwen3.5 family) that were unavailable at the time of the TMLR evaluation, so the comparison reflects both fine-tuning gains and generational model improvements.

\subsection{Comparison with Published Leaderboard Results}
\label{app:leaderboard}

To contextualize A3-Qwen3.5-9B within the broader landscape, Table~\ref{tab:leaderboard} compares our results against published results from official leaderboards across all five benchmarks. For each benchmark, we show the top proprietary model, the top open-weight model, and the best open-weight model under 10B parameters. Results are drawn from three sources: the WebArena leaderboard,\footnote{\url{https://docs.google.com/spreadsheets/d/1M801lEpBbKSNwP-vDBkC_pF7LdyGU1f_ufZb_NWNBZQ}} the VisualWebArena leaderboard (same spreadsheet, VWA tab), and the BrowserGym leaderboard.\footnote{\url{https://huggingface.co/spaces/ServiceNow/browsergym-leaderboard}}

\begin{table}[h]
\centering
\caption{A3-Qwen3.5-9B compared against official leaderboard results across all five benchmarks. For each benchmark, we show the top proprietary, open-weight, and sub-10B models. Our 9B SFT model is the best open-weight SFT result on full WebArena (nearly 2$\times$ Go-Browse) and exceeds GPT-4o on 4/5 benchmarks under the same GenericAgent protocol. Success rate (\%); $^\ddagger$ = custom agent pipeline; $^\dagger$ = test split only (leaderboard reports full benchmark). A3-Qwen3.5-9B and Qwen3.5-9B WA numbers are on the full 812-task benchmark.}
\label{tab:leaderboard}
\small
\setlength{\tabcolsep}{2.5pt}
\begin{tabular}{llcccccc}
\toprule
\textbf{Harness} & \textbf{Model} & \textbf{Params} & \textbf{WA} & \textbf{VWA} & \textbf{L1} & \textbf{L2} & \textbf{WoB} \\
\midrule
\multicolumn{8}{l}{\textit{Proprietary SOTA (from official leaderboards)}} \\
Custom$^\ddagger$ & OpAgent~\citep{Guo2026OpAgentOA} & -- & 71.6 & -- & -- & -- & -- \\
Custom$^\ddagger$ & SGV~\citep{Andrade2025LetsTI} & -- & -- & 54.0 & -- & -- & -- \\
GenericAgent & GPT-5~\citep{Chezelles2024TheBE} & -- & -- & -- & 79.1 & 69.4 & 71.5 \\
GenericAgent & Claude 4 Sonnet~\citep{Chezelles2024TheBE} & -- & -- & -- & 63.3 & 40.4 & 70.7 \\
\midrule
\multicolumn{8}{l}{\textit{Open-weight ($\geq$10B, from leaderboards)}} \\
Custom & TTI~\citep{Shen2025ThinkingVD} & 12B & 26.1 & -- & -- & -- & -- \\
Custom & Llama-3-70B + Search~\citep{Koh2024TreeSF} & 70B & -- & 16.7 & -- & -- & -- \\
GenericAgent & GPT-oss-120B~\citep{Chezelles2024TheBE} & 120B & -- & -- & 50.9 & 11.5 & 66.4 \\
\midrule
\multicolumn{8}{l}{\textit{Open-weight ($<$10B)}} \\
Custom & Go-Browse~\citep{Gandhi2025GoBrowseTW} & 7B & 21.7 & -- & -- & -- & -- \\
Custom & NNetNav~\citep{Murty2024NNetNavUL} & 8B & 18.8 & -- & -- & -- & -- \\
Custom & ViGoRL~\citep{Sarch2025GroundedRL} & 7B & -- & 11.2 & -- & -- & -- \\
\midrule
\multicolumn{8}{l}{\textit{This work (GenericAgent)}} \\
\textbf{GenericAgent} & \textbf{A3-Qwen3.5-9B} & \textbf{9B} & \textbf{42.1} & \textbf{33.7} & \textbf{51.5} & \textbf{9.7}$^\dagger$ & \textbf{69.0} \\
GenericAgent & Qwen3.5-9B (base) & 9B & 30.2 & 26.2 & 33.3 & 2.2$^\dagger$ & 63.2 \\
\bottomrule
\end{tabular}
\end{table}

\paragraph{Key observations.}

\begin{enumerate}[nosep,leftmargin=1.5em]
\item \textbf{Highest open-weight SFT on full WebArena.} A3-Qwen3.5-9B (42.1\% on the full 812-task benchmark) exceeds the previous best open-weight SFT result, Go-Browse (21.7\%), by 20.4pp. Go-Browse also uses AgentLab for evaluation, but with a different student model (Qwen-2.5-7B) and teacher (multiple models including Claude~3.7~Sonnet), so the comparison reflects differences in both the distillation pipeline and the student architecture.

\item \textbf{Gap to custom pipelines and next-generation models.} Custom multi-agent systems (OpAgent, 71.6\%) and next-generation proprietary models (GPT-5, Claude~4~Sonnet) substantially exceed our single-model SFT approach, particularly on enterprise tasks (WorkArena~L1/L2). These represent upper bounds for what more complex architectures or stronger base models can achieve.

\item \textbf{Few open-weight sub-10B baselines beyond WebArena.} While Go-Browse and NNetNav report on WebArena and ViGoRL on VWA, no sub-10B open-weight model reports results on WorkArena or MiniWoB, making A3-Qwen3.5-9B one of the first small open models evaluated across this full five-benchmark suite.
\end{enumerate}

\paragraph{Evaluation setup caveats.} OpAgent uses a multi-agent architecture, not a single-model setup. WebArena and VWA leaderboard entries use varied setups (SoM, screenshots, custom pipelines) that differ from GenericAgent. We omit WebRL~\citep{Qi2024WebRLTL} and WebAgent-R1~\citep{Wei2025WebAgentR1TW} because they report on WebArena-Lite (165 tasks), not the full 812-task benchmark.

\section{Ablation Studies}
\label{app:ablations}

\subsection{Judge Filtering Ablation}
\label{app:judge_ablation}

To measure the contribution of the Judge module, we trained Qwen3.5-9B on \emph{all} Pro (reduced thinking) trajectories without Judge filtering (both successful and failed trajectories included). This ``unfiltered'' variant uses 2,999 trajectories compared to 2,322 successful trajectories in the filtered version, providing 40\% more training data but without quality control.

The unfiltered model achieves \textbf{37.0\%} on WebArena (381 tasks), compared to \textbf{41.5\%} for the Judge-filtered model, a \textbf{4.5pp drop} (Table~\ref{tab:ablation_judge}). This confirms that the Judge module's quality filtering is beneficial: despite receiving substantially more training data, the student trained on noisy (unfiltered) trajectories underperforms the student trained on fewer but Judge-verified trajectories. The Judge's contribution is not merely removing clearly failed trajectories; it acts as a quality gate that improves the signal-to-noise ratio of the training data.

\begin{table}[h]
\centering
\caption{Judge filtering ablation on WebArena (381 tasks). Removing the Judge and training on all trajectories (including failed ones) reduces overall success rate by 4.5pp despite 40\% more training data. Per-site breakdowns show the largest drops on Map ($-$7.5pp), Shopping ($-$6.8pp), and GitLab ($-$6.6pp).}
\label{tab:ablation_judge}
\small
\setlength{\tabcolsep}{3.5pt}
\begin{tabular}{lccccccc}
\toprule
\textbf{Training Data} & \textbf{Trajs} & \textbf{All} & \textbf{Rdt} & \textbf{Git} & \textbf{Shop} & \textbf{ShAdm} & \textbf{Map} \\
\midrule
Judge-filtered (A3-Qwen3.5-9B) & 2,322 & \textbf{41.5} & 60.0 & \textbf{53.3} & \textbf{34.1} & \textbf{44.9} & \textbf{26.4} \\
Unfiltered (no Judge) & 2,999 & 37.0 & \textbf{62.2} & 46.7 & 27.3 & 42.3 & 18.9 \\
\midrule
\multicolumn{2}{l}{$\Delta$ (no Judge)} & $-$4.5 & +2.2 & \textcolor{red}{($-$6.6)} & \textcolor{red}{($-$6.8)} & \textcolor{red}{($-$2.6)} & \textcolor{red}{($-$7.5)} \\
\bottomrule
\end{tabular}
\end{table}

\subsection{Data Scaling Ablation}
\label{app:data_scaling}

To determine whether performance is saturating at our current data size or would benefit from more trajectories, we trained Qwen3.5-9B on three subsets of the Pro (reduced thinking) data: 285 trajectories (~2k steps), 715 (~5k steps), and 1,430 (~10k steps), compared to the full 2,322 (~16k steps).

\begin{table}[h]
\centering
\caption{Data scaling on WebArena (381 tasks). Performance increases with training data, from 32.0\% at 285 trajectories to 41.5\% at 2,322, with diminishing returns at larger scales.}
\label{tab:ablation_scaling}
\small
\setlength{\tabcolsep}{3.5pt}
\begin{tabular}{lcccccccc}
\toprule
\textbf{Variant} & \textbf{Trajs} & \textbf{Steps} & \textbf{All} & \textbf{Rdt} & \textbf{Git} & \textbf{Shop} & \textbf{ShAdm} & \textbf{Map} \\
\midrule
285-traj & 285 & 2,036 & 32.0 & 53.3 & 39.1 & 28.4 & 33.3 & 15.1 \\
715-traj & 715 & 5,060 & 37.0 & 57.8 & 46.7 & 27.3 & 42.3 & 22.6 \\
1430-traj & 1,430 & 9,999 & 40.2 & \textbf{62.2} & 46.7 & 35.2 & 46.2 & 20.8 \\
Full (A3-Qwen3.5-9B) & 2,322 & 16,353 & \textbf{41.5} & 60.0 & \textbf{53.3} & 35.2 & 46.2 & \textbf{26.4} \\
\bottomrule
\end{tabular}
\end{table}

\begin{figure}[h]
\centering
\includegraphics[width=\linewidth]{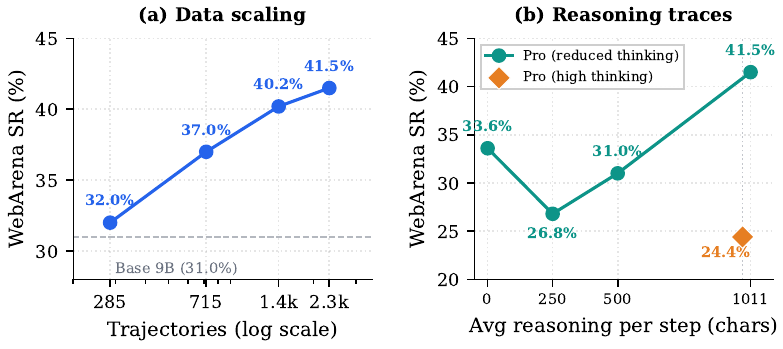}
\caption{(a)~WebArena SR increases with training data from 32.0\% (285 trajectories) to 41.5\% (2,322), with diminishing returns. (b)~Reasoning trace ablation: removing traces entirely ($-$7.9pp) is less harmful than truncating them ($-$10.5pp at 500 chars, $-$14.7pp at 250), suggesting incoherent mid-thought cutoffs degrade training signal more than the absence of reasoning.}
\label{fig:data_scaling}
\label{fig:reasoning_ablation}
\end{figure}

Performance increases from 32.0\% (285 trajectories) to 41.5\% (2,322 trajectories), with clear diminishing returns: +5.0pp from 285$\to$715, +3.2pp from 715$\to$1{,}430, and +1.3pp from 1{,}430$\to$2{,}322 (Figure~\ref{fig:data_scaling}). The smallest subset (285 trajectories) performs comparably to the base Qwen3.5-9B (32.0\% vs.\ 31.0\%), a difference of only 4 tasks on the 381-task benchmark, suggesting that very small amounts of distilled data provide marginal gains. Substantial improvements require at least several hundred trajectories.

\subsection{Reasoning Trace Ablation}
\label{app:reasoning_ablation}

Teacher trajectories contain structured reasoning traces in \texttt{<thought>} and \texttt{<think>} blocks (averaging 1,011 characters per block). To test whether these reasoning traces contribute to student performance, we trained three variants: truncated to 500 characters per block, truncated to 250 characters, and reasoning fully removed.

Reasoning traces are critical for student performance (Figure~\ref{fig:reasoning_ablation}). Removing them entirely drops WebArena SR from 41.5\% to 33.6\% ($-$7.9pp). Surprisingly, \emph{truncating} reasoning is worse than removing it entirely: Truncated-500 (31.0\%) and Truncated-250 (26.8\%) both underperform the no-reasoning variant. This suggests that truncated reasoning creates incoherent training signal: the student learns to produce reasoning that is cut off mid-thought, which may be worse than learning to act without explicit reasoning. Full reasoning traces, where the teacher's deliberation is preserved intact, produce the best student.

\paragraph{Teacher thinking budget.} Figure~\ref{fig:reasoning_ablation} also shows the result of training Qwen3.5-9B on data from Gemini~3~Pro with a high thinking budget (orange diamond). Despite the higher budget, the resulting reasoning traces are not actually longer than those from the reduced-thinking teacher. We hypothesize that with a larger thinking budget, the model performs more internal deliberation and produces more summarized outputs, externalizing less of its reasoning into the visible trace. The student, which can only learn from the visible text, receives a weaker training signal as a result (24.4\% vs.\ 41.5\%).

\subsection{Persona Ablation}
\label{app:persona_ablation}

To measure the contribution of the Persona Generator module, we trained Qwen3.5-9B on a subset of 600 persona-driven tasks (4,261 steps) subsampled from the full \websynth{} pipeline. The 600-task subset achieves 37.8\% on WebArena, a 3.7pp drop from the full 2,322-task model (41.5\%), retaining 91\% of the full pipeline's performance despite using only 26\% of the training data. This serves as the reference point for the hints ablation below: both the persona subset and the no-hints variant use matched 600-task training sets, so performance differences isolate the effect of hints rather than data quantity.

\subsection{Hints Ablation}
\label{app:hints_ablation}

To measure the contribution of evaluation hints to the Judge module, we re-judged all 2,999 Pro (reduced thinking) trajectories with the hints stripped from the Judge prompt. Without hints, the Judge's verdict flipped for 21.3\% of trajectories (632 of 2,968 re-judged), with 144 trajectories that the hint-assisted Judge marked as successful now judged as unsuccessful (false positives that hints helped catch). We trained Qwen3.5-9B on 600 tasks (4,195 steps) subsampled from the no-hints-filtered successful trajectories, matching the data size of the persona ablation for controlled comparison.

\begin{table}[h]
\centering
\caption{Hints ablation on WebArena (381 tasks). Removing hints from the Judge reduces downstream student performance by 2.4pp at matched data size, with the largest per-site drops on Shopping Admin ($-$6.4pp) and GitLab ($-$5.5pp).}
\label{tab:ablation_hints}
\small
\setlength{\tabcolsep}{3.5pt}
\begin{tabular}{lccccccc}
\toprule
\textbf{Training Data} & \textbf{Tasks} & \textbf{All} & \textbf{Rdt} & \textbf{Git} & \textbf{Shop} & \textbf{ShAdm} & \textbf{Map} \\
\midrule
Full pipeline (A3-Qwen3.5-9B) & 2,322 & \textbf{41.5} & \textbf{60.0} & \textbf{53.3} & 34.1 & \textbf{44.9} & \textbf{26.4} \\
With hints (subsampled) & 600 & 37.8 & 60.0 & 44.6 & 30.7 & 41.0 & 24.5 \\
No hints & 600 & 35.4 & 57.8 & 39.1 & \textbf{35.2} & 34.6 & 22.6 \\
\midrule
\multicolumn{2}{l}{$\Delta$ (hints contribution)} & $-$2.4 & \textcolor{red}{($-$2.2)} & \textcolor{red}{($-$5.5)} & +4.5 & \textcolor{red}{($-$6.4)} & \textcolor{red}{($-$1.9)} \\
\multicolumn{2}{l}{$\Delta$ (vs.\ full pipeline)} & $-$6.1 & \textcolor{red}{($-$2.2)} & \textcolor{red}{($-$14.2)} & +1.1 & \textcolor{red}{($-$10.3)} & \textcolor{red}{($-$3.8)} \\
\bottomrule
\end{tabular}
\end{table}

Removing hints drops WebArena SR from 37.8\% (with hints, matched data size) to 35.4\% ($-$2.4pp), or equivalently from 41.5\% (full pipeline) to 35.4\% ($-$6.1pp). The 2.4pp gap between the matched 600-task variants (with hints vs.\ without hints) isolates the hints' contribution to Judge accuracy: hints help the Judge correctly identify 144 additional false positives, improving the quality of the training data.

\section{Exploration Results}
\label{app:pipeline_results}
\label{app:exploration}

The exploration phase is a critical component of the Task Generator module. During exploration, an LLM agent is assigned a persona and navigates a web environment for up to 20 steps, building familiarity with the site's content, structure, and available functionalities. The exploration trajectory is stored in memory and later used as context for task generation.

\paragraph{Exploration Task Design.}
\label{app:exploration_design}

We create 1,500 exploration tasks distributed across the six WebArena environments (250 per site). Each task instructs the agent to explore a specific site while embodying an assigned persona, with the following prompt structure:

\begin{quote}
\small
``You have been instructed to explore the websites in order to familiarize yourself with their content and functionalities. When you are done, you should reply to the user with a message indicating that you are done exploring the websites: `I am done exploring the websites.' Make sure to explore for at least 10 steps before you stop.

You have been given the following persona: [persona description]''
\end{quote}

The exploration is considered successful if the agent produces the exact termination string (``I am done exploring the websites.'') after at least 10 steps of navigation. This simple success criterion ensures the agent actually explores rather than terminating prematurely.

\paragraph{Exploration Success Rates.}
\label{app:exploration_success}

Table~\ref{tab:exploration} reports exploration success rates for five models. Exploration success measures whether the model can navigate the environment and produce a valid termination signal; it does not evaluate the quality of exploration (which is implicitly reflected in downstream task quality).

\begin{table}[h]
\centering
\caption{Exploration success rates by model across 1,500 tasks (250 per site). Success requires the agent to explore for at least 10 steps and produce the termination string.}
\label{tab:exploration}
\small
\begin{tabular}{lccc}
\toprule
\textbf{Model} & \textbf{SR (\%)} & \textbf{Tasks Completed} & \textbf{Total Tasks} \\
\midrule
Gemini 3.1 Pro (reduced thinking) & 93.1 & 1,396 & 1,500 \\
Gemini 3 Flash & 88.3 & 1,325 & 1,500 \\
Gemini 3 Pro & 87.3 & 1,310 & 1,500 \\
Qwen3-VL-32B-Thinking & 74.7 & 1,121 & 1,500 \\
Qwen3-VL-8B-Thinking & 70.4 & 1,056 & 1,500 \\
\bottomrule
\end{tabular}
\end{table}

Proprietary models achieve substantially higher exploration success rates (87--93\%) compared to open-weight models (70--75\%). The primary failure mode for open-weight models is premature termination (stopping before 10 steps) or failure to produce the exact termination string. Gemini 3.1 Pro (reduced thinking) achieves the highest rate at 93.1\%, consistent with its strong performance as a teacher model on the downstream task generation pipeline.

\section{Framework and Implementation Details}
\label{app:framework_details}

This section details each module in the \ours{} framework.

\subsection{Vocabulary Mapping Across Pipelines}
\label{app:vocab_mapping}

Table~\ref{tab:vocab_mapping} maps the terminology used by each pipeline onto \ours{} module names. Despite different names, all four systems share the same underlying modular structure. InSTA's ``Task Proposer'' and Go-Browse's ``PageExplorer'' both fill the Task Generator role; NNetNav's ``LLM Task Labeler'' and Explorer's ``Task Summarizer'' do the same but retroactively. The Agent role is explicit in InSTA and Go-Browse, absent in NNetNav (explorations serve as data), and fused with the Task Refiner in Explorer. All four include a Judge under different names (ORM, Task Verifier, VLM-as-judge). No prior system produces evaluation Hints.

\begin{table}[h]
\centering
\caption{Vocabulary mapping across trajectory synthesis pipelines. ``--'' indicates the module is absent.}
\label{tab:vocab_mapping}
\small
\setlength{\tabcolsep}{3pt}
\begin{tabular}{lllll}
\toprule
\textbf{A3 Module} & \textbf{InSTA} & \textbf{NNetNav} & \textbf{Explorer} & \textbf{Go-Browse} \\
\midrule
Persona Gen. & -- & Persona DB & -- & -- \\
Exploration & Agent & LLM Explorer & Task Proposer & NavExplorer \\
Task Gen. & Task Proposer & Task Labeler & Refiner + Summ. & PageExplorer \\
Hints & -- & -- & -- & -- \\
Agent & Agent & -- & -- & Feas.Check + Solvers \\
Judge & LLM Judge & ORM & Task Verifier & VLM-as-judge \\
\bottomrule
\end{tabular}
\end{table}

\paragraph{Persona Generation.}
\label{app:persona_generation}

The Persona Generator module creates diverse user personas to drive task variety in the Task Generator. We generate 250 unique personas using an LLM, each characterized by a name, a set of professional skills (3 per persona), personal interests (3 per persona), and a detailed narrative description. The description includes the persona's professional background, expertise areas, personality traits, and how their interests connect to potential web activities. Each persona is assigned to all six environments, yielding 1,500 persona-environment pairs for exploration.

Persona diversity is critical for task coverage: a data scientist persona generates different tasks on a GitLab instance (e.g., ``Create a repository for my machine learning project'') than a graphic designer persona (e.g., ``Upload design assets to a new repository''). Each persona is assigned to multiple exploration sessions across different sites, ensuring that the intersection of persona characteristics and site affordances produces a combinatorially rich space of tasks.

\paragraph{Example personas.} Below we show three representative personas from the pool of 250:

\begin{enumerate}[nosep,leftmargin=1.5em]
\item \textbf{Alice Chen} (Data Scientist): Skills in Python Programming, Data Analysis, Machine Learning. Interests: Robotics, Hiking, Sci-Fi Novels. Alice specializes in transforming raw data into actionable intelligence; she spends weekends building custom robotics projects.

\item \textbf{Liam O'Connor} (Senior Graphic Designer): Skills in Graphic Design, Adobe Creative Suite, Typography. Interests: Analog Photography, Indie Music, Street Art. Liam merges digital precision with physical-world textures, shooting exclusively on film and collecting vinyl records.

\item \textbf{Dr.\ Fatima Al-Rashidi} (Biomedical Researcher): Skills in Bioinformatics, Statistical Analysis, Laboratory Techniques. Interests: Scientific Illustration, Mountaineering, Calligraphy. Fatima bridges computational biology with hands-on lab work, frequently presenting at international conferences.
\end{enumerate}

\subsection{Task Generation: Annotator Instructions}
\label{app:annotator_instructions}

The Task Generator receives \emph{annotator instructions} that guide the creation of high-quality, diverse tasks. These instructions are adapted from the original WebArena annotation guidelines~\citep{Zhou2023WebArenaAR} and specify three requirements:

\begin{enumerate}[nosep,leftmargin=1.5em]
\item \textbf{Abstract and high-level.} The intent should require multiple actions to complete, not merely one or two steps. For example, instead of ``click the science subreddit,'' annotators are encouraged to produce intents like ``post a greeting message on science subreddit,'' which requires navigation, form-filling, and submission.
\item \textbf{Creative.} Common tasks such as account creation are discouraged. Instead, annotators should add constraints (e.g., ``create a Reddit account identical to my GitLab one'') to produce unique intents.
\item \textbf{Template-based with variables.} Intents should be formulated as templates with replaceable elements marked as variables (e.g., \texttt{\{\{section\_name\}\}}, \texttt{\{\{topic\}\}}). Each template is instantiated with multiple variable assignments, producing diverse concrete tasks from a single template. For example, ``Browse the \texttt{\{\{section\_name\}\}} section to find a post containing \texttt{\{\{topic\}\}}'' can be instantiated as ``Browse the General section to find a post containing urban exploration'' or ``Browse the Off-Topic section to find a post containing analog photography.''
\end{enumerate}

These instructions are provided to the Task Generator LLM as part of its system prompt, along with the assigned persona and the exploration trajectory of the current environment.

\subsection{Example Synthesized Tasks}
\label{app:example_tasks}

We present five example tasks from \websynth{}, one per site, illustrating the diversity of synthesized intents, evaluation hints, and persona-driven grounding.

\paragraph{Reddit (persona: Liam O'Connor).}
\begin{itemize}[nosep,leftmargin=1.5em]
\item \textbf{Intent:} ``Browse the General section to find a post containing urban exploration and reply with a comment emphasizing the value of texture and grit.''
\item \textbf{Hint:} ``The agent must navigate to the specified section of the forum, identify a post that contains content related to urban exploration (e.g., by checking titles or images), and successfully submit a reply that explicitly mentions the design principle. Success is measured by the presence of the user's comment on the post page.''
\end{itemize}

\paragraph{GitLab (persona: Alice Chen).}
\begin{itemize}[nosep,leftmargin=1.5em]
\item \textbf{Intent:} ``Create a new public repository named `ml-data-pipeline' with a README file, then add an issue titled `Set up CI/CD pipeline' with the label `enhancement'.''
\item \textbf{Hint:} ``The agent should create the repository with the specified name, ensure it is public, and contains a README. Then navigate to the Issues section, create a new issue with the exact title, and apply the `enhancement' label. Success requires both the repository and issue to be visible.''
\end{itemize}

\paragraph{Shopping (persona: Dr.\ Fatima Al-Rashidi).}
\begin{itemize}[nosep,leftmargin=1.5em]
\item \textbf{Intent:} ``Find a laboratory notebook suitable for research documentation, add it to the shopping cart, and proceed to the checkout page.''
\item \textbf{Hint:} ``The agent should search for laboratory or research notebooks, select an appropriate product, add it to the cart, and navigate to the checkout page. The cart should contain at least one item and the agent should be on the checkout page at the end.''
\end{itemize}

\paragraph{Shopping Admin (persona: Alice Chen).}
\begin{itemize}[nosep,leftmargin=1.5em]
\item \textbf{Intent:} ``Navigate to the product catalog and change the price of the most expensive product in the `Electronics' category to \$99.99.''
\item \textbf{Hint:} ``The agent must access the product catalog, filter or navigate to the Electronics category, identify the most expensive product, edit its price to 99.99, and save the changes. The product's price should be updated to \$99.99 upon completion.''
\end{itemize}

\paragraph{Map (persona: Liam O'Connor).}
\begin{itemize}[nosep,leftmargin=1.5em]
\item \textbf{Intent:} ``Search for street art galleries near the city center and get directions from the nearest parking lot to the top-rated gallery.''
\item \textbf{Hint:} ``The agent should search for street art or art galleries, identify one near the center with good ratings, then search for a nearby parking lot and request directions between the two locations. Success requires the directions to be displayed on the map.''
\end{itemize}

\subsection{Judge Evaluation Protocol}
\label{app:judge_protocol}

The Judge module evaluates each trajectory to determine whether the agent successfully completed the synthesized task. Our Judge design builds on the evaluation protocol from AgentRewardBench~\citep{lu2025agentrewardbench}, which benchmarks automatic evaluators for web agents. We implement the Judge as an LLM that receives the full interaction record and produces structured evaluations.

\paragraph{Judge input.} For each trajectory, the Judge receives: (1) a system prompt defining its role as a web agent evaluator, (2) the task goal (intent), (3) the sequence of agent actions with URLs and reasoning at each step, (4) the final accessibility tree, and (5) the first and last screenshots of the interaction.

\paragraph{Evaluation questions.} The Judge answers four questions about the trajectory:

\begin{enumerate}[nosep,leftmargin=1.5em]
\item \textbf{Action looping} (\texttt{<loop>}): Did the agent loop through actions without making progress? (Yes/No)
\item \textbf{Side effects} (\texttt{<side>}): Did the agent perform unnecessary actions with unintended side effects? (Yes/No)
\item \textbf{Optimality} (\texttt{<optimal>}): Was the task performed optimally? (4-point scale: Complete Failure, Suboptimal, Somewhat Optimal, Completely Optimal)
\item \textbf{Success} (\texttt{<success>}): Was the task successfully completed? (Successful/Unsuccessful)
\end{enumerate}

Crucially, the success question is asked \emph{last} (after side effects, looping, and optimality), following an ``inverted'' ordering that forces the Judge to first consider potential issues before making the final success determination. This design reduces confirmation bias in success judgments.

\paragraph{Hint integration.} When evaluation hints are available (as in \websynth{}), they are appended to the task goal, providing the Judge with structured criteria for success evaluation. For example, the hint ``Success is measured by the presence of the user's comment on the post page'' gives the Judge an objective criterion that would be difficult to infer from the trajectory alone.

\paragraph{Parsing.} Judge responses are parsed by extracting content from the structured tags (\texttt{<success>}, \texttt{<side>}, \texttt{<loop>}, \texttt{<optimal>}). A trajectory is retained for training only if the \texttt{<success>} tag contains ``Successful.''

\subsection{Training Data Statistics}
\label{app:data_statistics}

Training data statistics per teacher configuration are in Table~\ref{tab:websynth_extended} (the ``Examples'' column). Each configuration produces trajectories on all six \websynth{} sites (500 tasks per site, 3,000 total). After Judge filtering, successful trajectories are converted to multi-turn observation-action pairs for supervised fine-tuning.

\paragraph{Data format.} Each training example consists of a multi-turn conversation: a system prompt specifying the agent's role and action format, a user message containing the current observation (accessibility tree, task goal, URL, and optionally a screenshot), and an assistant message containing the agent's reasoning and action. The assistant message follows a structured format: \texttt{<thought>}\ldots\texttt{</thought>} for high-level strategic reasoning, \texttt{<think>}\ldots\texttt{</think>} for step-by-step deliberation, and \texttt{<action>}\ldots\texttt{</action>} for the executable action command.

\paragraph{Action distribution.} In the Pro (reduced thinking) training data (16,353 examples), the action distribution is: \texttt{click} (65.3\%), \texttt{fill} (23.4\%), \texttt{send\_msg\_to\_user} (4.1\%), \texttt{select\_option} (2.5\%), \texttt{scroll} (1.5\%), \texttt{keyboard\_press} (1.5\%), \texttt{hover} (1.0\%), \texttt{goto} (0.4\%), \texttt{go\_back} (0.3\%), and other actions (0.2\%).

\paragraph{Response statistics.} The average assistant response length is 1,920 characters, with 100\% of responses containing explicit reasoning traces in \texttt{<thought>} blocks. The \texttt{<think>} tag (step-by-step reasoning) is present in 81.4\% of responses, with a median length of 459 characters when present. The \texttt{<thought>} tag (strategic reasoning) has a median length of 996 characters.

\subsection{Hyperparameters}
\label{app:hyperparameters}

Table~\ref{tab:hyperparameters} lists the complete training hyperparameters for our primary fine-tuned model, A3-Qwen3.5-9B.

\begin{table}[h]
\centering
\caption{Training hyperparameters for A3-Qwen3.5-9B.}
\label{tab:hyperparameters}
\small
\begin{tabular}{ll}
\toprule
\textbf{Hyperparameter} & \textbf{Value} \\
\midrule
Base model & Qwen/Qwen3.5-9B \\
Training data & Pro (reduced thinking) (16,353 examples) \\
\midrule
Learning rate & $1 \times 10^{-5}$ \\
LR schedule & Cosine with warmup \\
Warmup ratio & 0.03 \\
Weight decay & 0.0 \\
Max gradient norm & 0.3 \\
\midrule
Epochs & 2 \\
Batch size per GPU & 1 \\
Gradient accumulation steps & 4 \\
Effective batch size & 8 $\times$ 4 $\times$ 1 = 32 \\
\midrule
Max sequence length & 8,192 tokens \\
Precision & bfloat16 \\
Attention & Flash Attention 2 \\
Parallelism & FSDP (4--8 GPUs) \\
Optimizer & AdamW \\
\midrule
Checkpoint saving & Every 200 steps \\
Random seed & 0 \\
\bottomrule
\end{tabular}
\end{table}

\paragraph{Infrastructure.} Training is performed using Fully Sharded Data Parallelism~\citep{Zhao2023PyTorchFE} (FSDP). The primary model (A3-Qwen3.5-9B) was trained on 8 GPUs; ablation variants on 4 GPUs. We use the HuggingFace Transformers library with the TRL (Transformer Reinforcement Learning) framework for SFT.

\paragraph{Loss function.} We use the standard causal language modeling loss (cross-entropy) computed only on assistant tokens. System and user tokens are masked from the loss computation. This ensures the model learns to generate appropriate reasoning and actions conditioned on observations, without being trained to reproduce the observation format.

\section{Benchmark Details}
\label{app:benchmark_details}

\subsection{\websynth{} Details}
\label{app:websynth_details}

\paragraph{Task Distribution.}
\label{app:task_distribution}

\websynth{} consists of 3,000 synthesized tasks distributed evenly across six WebArena environments (500 tasks per site). Per-site success rates are in Table~\ref{tab:websynth_extended}.

\paragraph{Success rate variation.} The Pro (reduced thinking) teacher achieves the highest success rates on Wikipedia (85.4\%) and the lowest on Reddit (69.0\%). Wikipedia tasks tend to be more straightforward (editing pages, searching content), while Reddit tasks often involve multi-step interactions (finding specific posts, composing replies) that are more prone to failure.

\paragraph{Task categories.} Tasks span a wide range of web interaction types: content creation (posting, commenting, editing), information retrieval (searching, filtering, comparing), navigation (finding specific pages, following links), data management (creating records, updating fields), and transactional workflows (adding to cart, submitting forms). The persona-driven generation ensures coverage of diverse use cases within each site's affordances.

\paragraph{Relationship to WebArena Test Tasks.}
\label{app:websynth_vs_webarena}

\websynth{} tasks are synthesized on the same six WebArena environments used in the WebArena test set, but the synthesized task intents are entirely distinct from the 381 human-authored test tasks. The environments share the same web applications, data, and user accounts, meaning the model encounters familiar page layouts and interface elements during training. However, the specific goals, interaction sequences, and success criteria are novel. This makes WebArena an ``in-domain'' benchmark in terms of environment familiarity, but ``out-of-distribution'' with respect to task instructions. The consistent improvements on fully out-of-distribution benchmarks (VisualWebArena, WorkArena, MiniWoB) confirm that the learned skills generalize beyond environment-specific knowledge.

\paragraph{Comparison with Human-Authored Tasks.}
\label{app:websynth_vs_human}

The key differences between \websynth{} and the human-authored WebArena tasks are:

\begin{enumerate}[nosep,leftmargin=1.5em]
\item \textbf{Evaluation method.} WebArena uses programmatic evaluators (URL matching, string matching, HTML element checking) that require precise, hand-crafted evaluation functions per task. \websynth{} uses LLM-based judge evaluation with hints, enabling scalable evaluation without per-task programming.
\item \textbf{Task complexity.} WebArena tasks are carefully curated by human experts to span specific difficulty levels and interaction patterns. \websynth{} tasks are generated by the Task Generator LLM and may have different complexity distributions; some are simpler than WebArena tasks, while others attempt more creative interactions.
\item \textbf{Scale.} WebArena has 812 tasks (431 train + 381 test). \websynth{} produces 3,000 tasks per generation round, and the pipeline can be re-run to produce additional tasks by varying personas or regenerating explorations.
\item \textbf{Evaluation reliability.} Programmatic evaluators are deterministic but limited (they cannot evaluate open-ended outcomes). LLM judges are flexible but may introduce noise through false positives or false negatives. Hints mitigate this by providing the judge with structured evaluation criteria.
\end{enumerate}

\subsection{Evaluation Benchmark Summary}
\label{app:eval_benchmarks}

Table~\ref{tab:benchmark_summary} summarizes the key characteristics of all benchmarks used in our evaluation.

\begin{table}[h]
\centering
\caption{Summary of evaluation benchmarks.}
\label{tab:benchmark_summary}
\small
\begin{tabular}{lcccl}
\toprule
\textbf{Benchmark} & \textbf{Tasks} & \textbf{Steps} & \textbf{Domain} & \textbf{Relation to Training} \\
\midrule
WebArena & 812 & 30 & Self-hosted web apps & In-domain (same env) \\
VisualWebArena & 910$^\dagger$ & 30 & Visual web tasks & OOD (different tasks) \\
WorkArena L1 & 330 & 15 & ServiceNow (basic) & OOD (different env) \\
WorkArena++ L2 & 185 & 50 & ServiceNow (complex) & OOD (different env) \\
MiniWoB & 625 & 15 & Synthetic micro-tasks & OOD (different env) \\
\bottomrule
\end{tabular}
\end{table}

\paragraph{WebArena~\citep{Zhou2023WebArenaAR}.} A benchmark of realistic web tasks on six self-hosted web applications: a Reddit forum (Postmill), a GitLab instance, an e-commerce storefront (Magento), a store administration panel, a MediaWiki instance, and an OpenStreetMap deployment. Tasks require multi-step navigation, form-filling, and information retrieval. Evaluation combines URL matching, string matching, and programmatic HTML checks.

\paragraph{VisualWebArena~\citep{Koh2024VisualWebArenaEM}.} An extension of WebArena that includes visually grounded tasks requiring understanding of images, screenshots, and visual layouts. Tasks span three sites: classifieds, shopping, and Reddit. Evaluation is similar to WebArena.

\paragraph{WorkArena~\citep{Drouin2024WorkArenaHC} and WorkArena++~\citep{Boisvert2024WorkArenaTC}.} Benchmarks on ServiceNow, a widely-used enterprise software platform. WorkArena L1 tasks test basic operations (creating, filtering, sorting records; ordering catalog items; reading charts; navigation). WorkArena++ L2 tasks require multi-step compositional planning and reasoning across multiple ServiceNow modules, including dashboard interpretation, expense management, and multi-channel workflows. Evaluation uses ServiceNow API validation.

\paragraph{MiniWoB~\citep{Liu2018ReinforcementLO}.} A suite of simplified web interaction tasks testing atomic skills. Originally introduced by \citet{Shi2017WorldOB} and extended by \citet{Liu2018ReinforcementLO}, MiniWoB tasks include clicking buttons, filling forms, selecting options, and basic drag-and-drop operations. Tasks are rendered in a simplified HTML environment. Evaluation uses programmatic reward functions.

\section{Infrastructure and Evaluation Details}
\label{app:infrastructure}

\subsection{BrowserGym and AgentLab}
\label{app:browsergym}

All experiments use BrowserGym~\citep{Chezelles2024TheBE} as the environment interface and AgentLab as the evaluation framework. BrowserGym provides a unified API for browser-based web agent benchmarks, wrapping each benchmark as an OpenAI Gymnasium environment.

\paragraph{Observation space.} At each step, the agent receives: (1) the current page URL, (2) an accessibility tree (AXTree) representation of the page content, which provides a hierarchical, text-based view of all interactive elements with unique bid (browsing ID) identifiers, (3) a screenshot of the current page, and (4) the task goal and the agent's previous action. All multimodal models (including Qwen3.5-9B) receive both the AXTree and the screenshot at each step.

\paragraph{Action space.} The agent produces actions in a structured text format. Supported actions include: \texttt{click(bid)} to click an element, \texttt{fill(bid, text)} to type text into a field, \texttt{select\_option(bid, option)} to choose a dropdown option, \texttt{scroll(x, y)} to scroll the page, \texttt{keyboard\_press(key)} for keyboard shortcuts, \texttt{hover(bid)} for mouse hover, \texttt{goto(url)} for URL navigation, \texttt{go\_back()} for browser back, and \texttt{send\_msg\_to\_user(text)} to communicate with the user.

\paragraph{Evaluation protocol.} Each benchmark provides its own success evaluation: WebArena and VisualWebArena use a combination of URL matching, string matching, and programmatic HTML checks; WorkArena uses ServiceNow API validation; MiniWoB uses programmatic reward functions. For \websynth{}, we use LLM-based judge evaluation with hints (described in \secref{app:judge_protocol}).

\paragraph{Error handling and environment reliability.} Web-based evaluation environments are inherently fragile: the original BrowserGym evaluation framework reports significant error rates across benchmarks due to browser crashes, network timeouts, and environment state corruption~\citep{Chezelles2024TheBE}. We adopt several mitigations to reduce these errors: each task is retried up to 10 times on failure, the WebArena instance is fully reset between evaluation runs, and we use isolated browser contexts per task. Despite these measures, a residual error rate persists due to issues outside model control.

We categorize the remaining errors into three types:
\begin{itemize}[nosep,leftmargin=1.5em]
\item \textbf{Environment bugs:} Some WorkArena L2 tasks fail consistently across all models with ``Catalog item not found'' errors, indicating missing catalog items in the ServiceNow instance pool. Four such tasks (e.g., \texttt{dashboard-retrieve-catalog-and-*-order-loaner-laptop-l2}) fail for every model we evaluate, confirming this is an environment setup issue rather than a model limitation.
\item \textbf{Browser/network failures:} Playwright browser crashes (``Target crashed'', ``Target page, context or browser has been closed''), navigation timeouts on unresponsive pages, and transient connection errors between the evaluation host and self-hosted web instances. These affect 1--2 tasks per model on WebArena and are not systematic.
\item \textbf{Context length exceeded:} For smaller models (e.g., Qwen3.5-4B), some VisualWebArena tasks with large pages exceed the model's context window, causing API errors. This affects up to 11 tasks for the 4B model but does not occur for 9B+ models.
\end{itemize}

All reported success rates include these failed tasks as failures; we do not exclude them from the denominator.

\subsection{Compute Resources}
\label{app:compute}

\paragraph{Training.} Fine-tuning uses FSDP across 4--8 GPUs (A100 80GB or H200 141GB, depending on cluster availability). A single training run of Qwen3.5-9B for 2 epochs takes approximately 12--16 hours on 8 GPUs.

\paragraph{Inference.} Model serving for evaluation uses vLLM~\citep{Kwon2023EfficientMM} with tensor parallelism across 2 GPUs (A100 or H100). We use the enforce-eager mode (disabling CUDA graphs) for reliability with fine-tuned checkpoints. Models are served with the OpenAI-compatible API endpoint.

\paragraph{Evaluation.} Web browser interactions are driven by Playwright running in headless Chromium. Each evaluation task runs in an isolated browser context. Evaluation of the full suite (WebArena + VisualWebArena + WorkArena L1 + WorkArena++ L2 + MiniWoB) for a single model takes approximately 24--48 hours depending on the model's inference speed and the step limits per benchmark.

\subsection{Token Budget}
\label{app:token_budget}

All models use a unified token budget for fair comparison:

\begin{itemize}[nosep,leftmargin=1.5em]
\item \textbf{Maximum total tokens:} 65,536
\item \textbf{Maximum prompt tokens:} 57,344
\item \textbf{Maximum new tokens (generation):} 8,192
\end{itemize}

When the observation (primarily the AXTree) exceeds the prompt token limit, it is truncated. The generous prompt budget (57K tokens) accommodates the large AXTrees produced by complex web pages, which can exceed 30K tokens for pages with many interactive elements.

\label{app:step_limits}

%% file: references.bib
@Article{Trabucco2025TowardsIT,
 author = {Brandon Trabucco and Gunnar Sigurdsson and Robinson Piramuthu and Ruslan Salakhutdinov},
 booktitle = {arXiv.org},
 journal = {ArXiv},
 title = {Towards Internet-Scale Training For Agents},
 volume = {abs/2502.06776},
 year = {2025},
 url = {https://arxiv.org/abs/2502.06776}
}

@Inproceedings{Murty2024NNetNavUL,
 author = {Shikhar Murty and Dzmitry Bahdanau and Christopher D. Manning},
 title = {NNetNav: Unsupervised Learning of Browser Agents Through Environment Interaction in the Wild},
 year = {2024},
 url = {https://arxiv.org/abs/2410.02907}
}

@Article{Zhou2023WebArenaAR,
 author = {Shuyan Zhou and Frank F. Xu and Hao Zhu and Xuhui Zhou and Robert Lo and Abishek Sridhar and Xianyi Cheng and Yonatan Bisk and Daniel Fried and Uri Alon and Graham Neubig},
 booktitle = {International Conference on Learning Representations},
 journal = {ArXiv},
 title = {WebArena: A Realistic Web Environment for Building Autonomous Agents},
 volume = {abs/2307.13854},
 year = {2023},
 url = {https://arxiv.org/abs/2307.13854}
}

@Article{Chezelles2024TheBE,
 author = {Thibault Le Sellier de Chezelles and Maxime Gasse and Alexandre Lacoste and Alexandre Drouin and Massimo Caccia and L{\'e}o Boisvert and Megh Thakkar and Tom Marty and Rim Assouel and Sahar Omidi Shayegan and Lawrence Jang and Xing Han L{\`u} and Ori Yoran and Dehan Kong and Frank F. Xu and Siva Reddy and Quentin Cappart and Graham Neubig and Ruslan Salakhutdinov and Nicolas Chapados},
 booktitle = {Trans. Mach. Learn. Res.},
 journal = {ArXiv},
 title = {The BrowserGym Ecosystem for Web Agent Research},
 volume = {abs/2412.05467},
 year = {2024},
 url = {https://arxiv.org/abs/2412.05467}
}

@Article{Koh2024VisualWebArenaEM,
 author = {Jing Yu Koh and Robert Lo and Lawrence Jang and Vikram Duvvur and Ming Chong Lim and Po-Yu Huang and Graham Neubig and Shuyan Zhou and Ruslan Salakhutdinov and Daniel Fried},
 booktitle = {Annual Meeting of the Association for Computational Linguistics},
 journal = {ArXiv},
 title = {VisualWebArena: Evaluating Multimodal Agents on Realistic Visual Web Tasks},
 volume = {abs/2401.13649},
 year = {2024},
 url = {https://arxiv.org/abs/2401.13649}
}

@Article{Drouin2024WorkArenaHC,
 author = {Alexandre Drouin and Maxime Gasse and Massimo Caccia and I. Laradji and Manuel Del Verme and Tom Marty and L{\'e}o Boisvert and Megh Thakkar and Quentin Cappart and David V{\'a}zquez and Nicolas Chapados and Alexandre Lacoste},
 booktitle = {International Conference on Machine Learning},
 journal = {ArXiv},
 title = {WorkArena: How Capable Are Web Agents at Solving Common Knowledge Work Tasks?},
 volume = {abs/2403.07718},
 year = {2024},
 url = {https://arxiv.org/abs/2403.07718}
}

@Article{Yoran2024AssistantBenchCW,
 author = {Ori Yoran and S. Amouyal and Chaitanya Malaviya and Ben Bogin and Ofir Press and Jonathan Berant},
 booktitle = {Conference on Empirical Methods in Natural Language Processing},
 pages = {8938-8968},
 title = {AssistantBench: Can Web Agents Solve Realistic and Time-Consuming Tasks?},
 year = {2024},
 url = {https://arxiv.org/abs/2407.15711}
}

@Article{Liu2018ReinforcementLO,
 author = {E. Liu and Kelvin Guu and Panupong Pasupat and Tianlin Shi and Percy Liang},
 booktitle = {International Conference on Learning Representations},
 journal = {ArXiv},
 title = {Reinforcement Learning on Web Interfaces Using Workflow-Guided Exploration},
 volume = {abs/1802.08802},
 year = {2018},
 url = {https://arxiv.org/abs/1802.08802}
}

@Article{Xu2024AgentTrekAT,
 author = {Yiheng Xu and Dunjie Lu and Zhennan Shen and Junli Wang and Zekun Wang and Yuchen Mao and Caiming Xiong and Tao Yu},
 booktitle = {International Conference on Learning Representations},
 journal = {ArXiv},
 title = {AgentTrek: Agent Trajectory Synthesis via Guiding Replay with Web Tutorials},
 volume = {abs/2412.09605},
 year = {2024},
 url = {https://arxiv.org/abs/2412.09605}
}

@Article{Bai2024DigiRLTI,
 author = {Hao Bai and Yifei Zhou and M. Cemri and Jiayi Pan and Alane Suhr and Sergey Levine and Aviral Kumar},
 booktitle = {Neural Information Processing Systems},
 journal = {ArXiv},
 title = {DigiRL: Training In-The-Wild Device-Control Agents with Autonomous Reinforcement Learning},
 volume = {abs/2406.11896},
 year = {2024},
 url = {https://arxiv.org/abs/2406.11896}
}

@Article{Wang2024AgentWM,
 author = {Z. Wang and Jiayuan Mao and Daniel Fried and Graham Neubig},
 booktitle = {International Conference on Machine Learning},
 journal = {ArXiv},
 title = {Agent Workflow Memory},
 volume = {abs/2409.07429},
 year = {2024},
 url = {https://arxiv.org/abs/2409.07429}
}

@Article{He2024OpenWebVoyagerBM,
 author = {Hongliang He and Wenlin Yao and Kaixin Ma and Wenhao Yu and Hongming Zhang and Tianqing Fang and Zhenzhong Lan and Dong Yu},
 booktitle = {Annual Meeting of the Association for Computational Linguistics},
 pages = {27545-27564},
 title = {OpenWebVoyager: Building Multimodal Web Agents via Iterative Real-World Exploration, Feedback and Optimization},
 year = {2024},
 url = {https://arxiv.org/abs/2410.19609}
}

@Inproceedings{Zelikman2022STaRBR,
 author = {E. Zelikman and Yuhuai Wu and Noah D. Goodman},
 title = {STaR: Bootstrapping Reasoning With Reasoning},
 year = {2022},
 url = {https://arxiv.org/abs/2203.14465}
}

@Article{Shi2017WorldOB,
 author = {Tianlin Shi and A. Karpathy and Linxi (Jim) Fan and J. Hern{\'a}ndez and Percy Liang},
 booktitle = {International Conference on Machine Learning},
 pages = {3135-3144},
 title = {World of Bits: An Open-Domain Platform for Web-Based Agents},
 year = {2017},
 url = {https://www.semanticscholar.org/paper/298a55ddc9777e39c5bad92a750827e1cae98ac1}
}

@Article{Bai2025Qwen25VLTR,
 author = {Shuai Bai and Keqin Chen and Xuejing Liu and Jialin Wang and Wenbin Ge and Sibo Song and Kai Dang and Peng Wang and Shijie Wang and Jun Tang and Humen Zhong and Yuanzhi Zhu and Mingkun Yang and Zhaohai Li and Jianqiang Wan and Pengfei Wang and Wei Ding and Zheren Fu and Yiheng Xu and Jiabo Ye and Xi Zhang and Tianbao Xie and Zesen Cheng and Hang Zhang and Zhibo Yang and Haiyang Xu and Junyang Lin},
 booktitle = {arXiv.org},
 journal = {ArXiv},
 title = {Qwen2.5-VL Technical Report},
 volume = {abs/2502.13923},
 year = {2025},
 url = {https://arxiv.org/abs/2502.13923}
}

@Article{Boisvert2024WorkArenaTC,
 author = {L{\'e}o Boisvert and Megh Thakkar and Maxime Gasse and Massimo Caccia and Thibault Le Sellier de Chezelles and Quentin Cappart and Nicolas Chapados and Alexandre Lacoste and Alexandre Drouin},
 booktitle = {Neural Information Processing Systems},
 journal = {ArXiv},
 title = {WorkArena++: Towards Compositional Planning and Reasoning-based Common Knowledge Work Tasks},
 volume = {abs/2407.05291},
 year = {2024},
 url = {https://arxiv.org/abs/2407.05291}
}

@Article{Wang2022SelfInstructAL,
 author = {Yizhong Wang and Yeganeh Kordi and Swaroop Mishra and Alisa Liu and Noah A. Smith and Daniel Khashabi and Hannaneh Hajishirzi},
 booktitle = {Annual Meeting of the Association for Computational Linguistics},
 pages = {13484-13508},
 title = {Self-Instruct: Aligning Language Models with Self-Generated Instructions},
 year = {2022},
 url = {https://arxiv.org/abs/2212.10560}
}

@Article{Xu2023WizardLMEL,
 author = {Can Xu and Qingfeng Sun and Kai Zheng and Xiubo Geng and Pu Zhao and Jiazhan Feng and Chongyang Tao and Daxin Jiang},
 booktitle = {International Conference on Learning Representations},
 title = {WizardLM: Empowering Large Pre-Trained Language Models to Follow Complex Instructions},
 year = {2023},
 url = {https://arxiv.org/abs/2304.12244}
}

@Article{Xu2024MagpieAD,
 author = {Zhangchen Xu and Fengqing Jiang and Luyao Niu and Yuntian Deng and R. Poovendran and Yejin Choi and Bill Yuchen Lin},
 booktitle = {International Conference on Learning Representations},
 journal = {ArXiv},
 title = {Magpie: Alignment Data Synthesis from Scratch by Prompting Aligned LLMs with Nothing},
 volume = {abs/2406.08464},
 year = {2024},
 url = {https://arxiv.org/abs/2406.08464}
}

@Article{Chan2024ScalingSD,
 author = {Tao Ge and Xin Chan and Xiaoyang Wang and Dian Yu and Haitao Mi and Dong Yu},
 booktitle = {arXiv.org},
 journal = {ArXiv},
 title = {Scaling Synthetic Data Creation with 1,000,000,000 Personas},
 volume = {abs/2406.20094},
 year = {2024},
 url = {https://arxiv.org/abs/2406.20094}
}

@Article{Hsieh2023DistillingSO,
 author = {Cheng-Yu Hsieh and Chun-Liang Li and Chih-Kuan Yeh and Hootan Nakhost and Yasuhisa Fujii and Alexander J. Ratner and Ranjay Krishna and Chen-Yu Lee and Tomas Pfister},
 booktitle = {Annual Meeting of the Association for Computational Linguistics},
 journal = {ArXiv},
 title = {Distilling Step-by-Step! Outperforming Larger Language Models with Less Training Data and Smaller Model Sizes},
 volume = {abs/2305.02301},
 year = {2023},
 url = {https://arxiv.org/abs/2305.02301}
}

@Article{Mukherjee2023OrcaPL,
 author = {Subhabrata Mukherjee and Arindam Mitra and Ganesh Jawahar and Sahaj Agarwal and Hamid Palangi and A. Awadallah},
 booktitle = {arXiv.org},
 journal = {ArXiv},
 title = {Orca: Progressive Learning from Complex Explanation Traces of GPT-4},
 volume = {abs/2306.02707},
 year = {2023},
 url = {https://arxiv.org/abs/2306.02707}
}

@Article{Zhou2023LIMALI,
 author = {Chunting Zhou and Pengfei Liu and Puxin Xu and Srini Iyer and Jiao Sun and Yuning Mao and Xuezhe Ma and Avia Efrat and Ping Yu and L. Yu and Susan Zhang and Gargi Ghosh and M. Lewis and Luke Zettlemoyer and Omer Levy},
 booktitle = {Neural Information Processing Systems},
 journal = {ArXiv},
 title = {LIMA: Less Is More for Alignment},
 volume = {abs/2305.11206},
 year = {2023},
 url = {https://arxiv.org/abs/2305.11206}
}

@Article{Ouyang2022TrainingLM,
 author = {Long Ouyang and Jeff Wu and Xu Jiang and Diogo Almeida and Carroll L. Wainwright and Pamela Mishkin and Chong Zhang and S. Agarwal and Katarina Slama and Alex Ray and John Schulman and Jacob Hilton and Fraser Kelton and Luke E. Miller and M. Simens and Amanda Askell and Peter Welinder and P. Christiano and Jan Leike and Ryan J. Lowe},
 booktitle = {Neural Information Processing Systems},
 journal = {ArXiv},
 title = {Training language models to follow instructions with human feedback},
 volume = {abs/2203.02155},
 year = {2022},
 url = {https://arxiv.org/abs/2203.02155}
}

@Article{Zheng2023JudgingLW,
 author = {Lianmin Zheng and Wei-Lin Chiang and Ying Sheng and Siyuan Zhuang and Zhanghao Wu and Yonghao Zhuang and Zi Lin and Zhuohan Li and Dacheng Li and E. Xing and Haotong Zhang and Joseph E. Gonzalez and Ion Stoica},
 booktitle = {Neural Information Processing Systems},
 journal = {ArXiv},
 title = {Judging LLM-as-a-judge with MT-Bench and Chatbot Arena},
 volume = {abs/2306.05685},
 year = {2023},
 url = {https://arxiv.org/abs/2306.05685}
}

@Article{lu2025agentrewardbench,
 author = {Xing Han L{\`u} and Amirhossein Kazemnejad and Nicholas Meade and Arkil Patel and Dongchan Shin and Alejandra Zambrano and Karolina Sta{\'n}czak and Peter Shaw and Christopher Pal and Siva Reddy},
 booktitle = {arXiv.org},
 journal = {ArXiv},
 title = {AgentRewardBench: Evaluating Automatic Evaluations of Web Agent Trajectories},
 volume = {abs/2504.08942},
 year = {2025},
 url = {https://arxiv.org/abs/2504.08942}
}

@Article{Yao2022ReActSR,
 author = {Shunyu Yao and Jeffrey Zhao and Dian Yu and Nan Du and Izhak Shafran and Karthik Narasimhan and Yuan Cao},
 booktitle = {International Conference on Learning Representations},
 journal = {ArXiv},
 title = {ReAct: Synergizing Reasoning and Acting in Language Models},
 volume = {abs/2210.03629},
 year = {2022},
 url = {https://arxiv.org/abs/2210.03629}
}

@Article{Lu2024WebLINXRW,
 author = {Xing Han L{\`u} and Zden{\v{e}}k Kasner and Siva Reddy},
 booktitle = {International Conference on Machine Learning},
 pages = {33007-33056},
 title = {WebLINX: Real-World Website Navigation with Multi-Turn Dialogue},
 year = {2024},
 url = {https://arxiv.org/abs/2402.05930}
}

@Article{Zheng2024GPT4VisionIA,
 author = {Boyuan Zheng and Boyu Gou and Jihyung Kil and Huan Sun and Yu Su},
 booktitle = {International Conference on Machine Learning},
 pages = {61349-61385},
 title = {GPT-4V(ision) is a Generalist Web Agent, if Grounded},
 year = {2024},
 url = {https://arxiv.org/abs/2401.01614}
}

@Article{Deng2023Mind2WebTA,
 author = {Xiang Deng and Yu Gu and Boyuan Zheng and Shijie Chen and Samuel Stevens and Boshi Wang and Huan Sun and Yu Su},
 booktitle = {Neural Information Processing Systems},
 journal = {ArXiv},
 title = {Mind2Web: Towards a Generalist Agent for the Web},
 volume = {abs/2306.06070},
 year = {2023},
 url = {https://arxiv.org/abs/2306.06070}
}

@Article{Yao2022WebShopTS,
 author = {Shunyu Yao and Howard Chen and John Yang and Karthik Narasimhan},
 booktitle = {Neural Information Processing Systems},
 journal = {ArXiv},
 title = {WebShop: Towards Scalable Real-World Web Interaction with Grounded Language Agents},
 volume = {abs/2207.01206},
 year = {2022},
 url = {https://arxiv.org/abs/2207.01206}
}

@Article{Xie2024OSWorldBM,
 author = {Tianbao Xie and Danyang Zhang and Jixuan Chen and Xiaochuan Li and Siheng Zhao and Ruisheng Cao and T. Hua and Zhoujun Cheng and Dongchan Shin and Fangyu Lei and Yitao Liu and Yiheng Xu and Shuyan Zhou and Silvio Savarese and Caiming Xiong and Victor Zhong and Tao Yu},
 booktitle = {Neural Information Processing Systems},
 journal = {ArXiv},
 title = {OSWorld: Benchmarking Multimodal Agents for Open-Ended Tasks in Real Computer Environments},
 volume = {abs/2404.07972},
 year = {2024},
 url = {https://arxiv.org/abs/2404.07972}
}

@Article{Wang2023ASO,
 author = {Lei Wang and Chengbang Ma and Xueyang Feng and Zeyu Zhang and Hao-ran Yang and Jingsen Zhang and Zhi-Yang Chen and Jiakai Tang and Xu Chen and Yankai Lin and Wayne Xin Zhao and Zhewei Wei and Ji-rong Wen},
 booktitle = {Frontiers of Computer Science},
 journal = {Frontiers of Computer Science},
 title = {A survey on large language model based autonomous agents},
 volume = {18},
 year = {2023},
 url = {https://arxiv.org/abs/2308.11432}
}

@Article{Putta2024AgentQA,
 author = {Pranav Putta and Edmund Mills and Naman Garg and S. Motwani and Chelsea Finn and Divyansh Garg and Rafael Rafailov},
 booktitle = {arXiv.org},
 journal = {ArXiv},
 title = {Agent Q: Advanced Reasoning and Learning for Autonomous AI Agents},
 volume = {abs/2408.07199},
 year = {2024},
 url = {https://arxiv.org/abs/2408.07199}
}

@Inproceedings{Sodhi2023StePSL,
 author = {Paloma Sodhi and S. Branavan and Yoav Artzi and Ryan Mcdonald},
 title = {SteP: Stacked LLM Policies for Web Actions},
 year = {2023},
 url = {https://arxiv.org/abs/2310.03720}
}

@Article{Gur2023ARW,
 author = {Izzeddin Gur and Hiroki Furuta and Austin Huang and Mustafa Safdari and Yutaka Matsuo and D. Eck and Aleksandra Faust},
 booktitle = {International Conference on Learning Representations},
 journal = {ArXiv},
 title = {A Real-World WebAgent with Planning, Long Context Understanding, and Program Synthesis},
 volume = {abs/2307.12856},
 year = {2023},
 url = {https://arxiv.org/abs/2307.12856}
}

@Article{Hong2023CogAgentAV,
 author = {Wenyi Hong and Weihan Wang and Qingsong Lv and Jiazheng Xu and Wenmeng Yu and Junhui Ji and Yan Wang and Zihan Wang and Yuxiao Dong and Ming Ding and Jie Tang},
 booktitle = {Computer Vision and Pattern Recognition},
 journal = {2024 IEEE/CVF Conference on Computer Vision and Pattern Recognition (CVPR)},
 pages = {14281-14290},
 title = {CogAgent: A Visual Language Model for GUI Agents},
 year = {2023},
 url = {https://arxiv.org/abs/2312.08914}
}

@Article{Liu2023AgentBenchEL,
 author = {Xiao Liu and Hao Yu and Hanchen Zhang and Yifan Xu and Xuanyu Lei and Hanyu Lai and Yu Gu and Yuxian Gu and Han Ding and Kai Men and Kejuan Yang and Shudan Zhang and Xiang Deng and Aohan Zeng and Zhengxiao Du and Chenhui Zhang and Shengqi Shen and Tianjun Zhang and Sheng Shen and Yu Su and Huan Sun and Minlie Huang and Yuxiao Dong and Jie Tang},
 booktitle = {International Conference on Learning Representations},
 journal = {ArXiv},
 title = {AgentBench: Evaluating LLMs as Agents},
 volume = {abs/2308.03688},
 year = {2023},
 url = {https://arxiv.org/abs/2308.03688}
}

@Article{Rawles2024AndroidWorldAD,
 author = {Christopher Rawles and Sarah Clinckemaillie and Yifan Chang and J. Waltz and G. Lau and Marybeth Fair and Alice Li and Will Bishop and Wei Li and Folawiyo Campbell-Ajala and Daniel Toyama and Robert Berry and Divya Tyamagundlu and Timothy P. Lillicrap and O. Riva},
 booktitle = {International Conference on Learning Representations},
 journal = {ArXiv},
 title = {AndroidWorld: A Dynamic Benchmarking Environment for Autonomous Agents},
 volume = {abs/2405.14573},
 year = {2024},
 url = {https://arxiv.org/abs/2405.14573}
}

@Article{Qi2024WebRLTL,
 author = {Zehan Qi and Xiao Liu and Iat Long Iong and Hanyu Lai and Xueqiao Sun and Xinyue Yang and Jiadai Sun and Yu Yang and Shuntian Yao and Tianjie Zhang and Wei Xu and Jie Tang and Yuxiao Dong},
 booktitle = {arXiv.org},
 journal = {ArXiv},
 title = {WebRL: Training LLM Web Agents via Self-Evolving Online Curriculum Reinforcement Learning},
 volume = {abs/2411.02337},
 year = {2024},
 url = {https://arxiv.org/abs/2411.02337}
}

@InProceedings{Pahuja2025ExplorerSE,
 author = {Vardaan Pahuja and Yadong Lu and Corby Rosset and Boyu Gou and Arindam Mitra and Spencer Whitehead and Yu Su and Ahmed Awadallah},
 booktitle = {Annual Meeting of the Association for Computational Linguistics},
 pages = {6300-6323},
 title = {Explorer: Scaling Exploration-driven Web Trajectory Synthesis for Multimodal Web Agents},
 year = {2025},
 url = {https://arxiv.org/abs/2502.11357}
}

@Article{Xu2024TheAgentCompanyBL,
 author = {Frank F. Xu and Yufan Song and Boxuan Li and Yuxuan Tang and Kritanjali Jain and Meng Bao and Z. Wang and Xuhui Zhou and Zhitong Guo and Murong Cao and Ming-Hsuan Yang and Hao Lu and Amaad Martin and Zhe Su and L. Maben and Raj Mehta and Wayne Chi and Lawrence Jang and Yiqing Xie and Shuyan Zhou and Graham Neubig},
 booktitle = {arXiv.org},
 journal = {ArXiv},
 title = {TheAgentCompany: Benchmarking LLM Agents on Consequential Real World Tasks},
 volume = {abs/2412.14161},
 year = {2024},
 url = {https://arxiv.org/abs/2412.14161}
}

@Article{Pan2024WebCanvasBW,
 author = {Yichen Pan and Dehan Kong and Sida Zhou and Cheng Cui and Yifei Leng and Bingqian Jiang and Hangyu Liu and Yanyi Shang and Shuyan Zhou and Tongshuang Wu and Zhengyang Wu},
 booktitle = {arXiv.org},
 journal = {ArXiv},
 title = {WebCanvas: Benchmarking Web Agents in Online Environments},
 volume = {abs/2406.12373},
 year = {2024},
 url = {https://arxiv.org/abs/2406.12373}
}

@Article{Zhou2023LanguageAT,
 author = {Andy Zhou and Kai Yan and Michal Shlapentokh-Rothman and Haohan Wang and Yu-Xiong Wang},
 booktitle = {International Conference on Machine Learning},
 journal = {ArXiv},
 title = {Language Agent Tree Search Unifies Reasoning Acting and Planning in Language Models},
 volume = {abs/2310.04406},
 year = {2023},
 url = {https://arxiv.org/abs/2310.04406}
}

@Article{He2024WebVoyagerBA,
 author = {Hongliang He and Wenlin Yao and Kaixin Ma and Wenhao Yu and Yong Dai and Hongming Zhang and Zhenzhong Lan and Dong Yu},
 booktitle = {Annual Meeting of the Association for Computational Linguistics},
 journal = {ArXiv},
 title = {WebVoyager: Building an End-to-End Web Agent with Large Multimodal Models},
 volume = {abs/2401.13919},
 year = {2024},
 url = {https://arxiv.org/abs/2401.13919}
}

@Article{Furuta2023MultimodalWN,
 author = {Hiroki Furuta and Ofir Nachum and Kuang-Huei Lee and Yutaka Matsuo and S. Gu and Izzeddin Gur},
 booktitle = {International Conference on Learning Representations},
 journal = {ArXiv},
 title = {Multimodal Web Navigation with Instruction-Finetuned Foundation Models},
 volume = {abs/2305.11854},
 year = {2023},
 url = {https://arxiv.org/abs/2305.11854}
}

@Book{Lai2024AutoWebGLMAL,
 author = {Hanyu Lai and Xiao Liu and Iat Long Iong and Shuntian Yao and Yuxuan Chen and Pengbo Shen and Hao Yu and Hanchen Zhang and Xiaohan Zhang and Yuxiao Dong and Jie Tang},
 booktitle = {Knowledge Discovery and Data Mining},
 journal = {Proceedings of the 30th ACM SIGKDD Conference on Knowledge Discovery and Data Mining},
 title = {AutoWebGLM: A Large Language Model-based Web Navigating Agent},
 year = {2024},
 url = {https://arxiv.org/abs/2404.03648}
}

@Article{Zeng2023AgentTuningEG,
 author = {Aohan Zeng and Mingdao Liu and Rui Lu and Bowen Wang and Xiao Liu and Yuxiao Dong and Jie Tang},
 booktitle = {Annual Meeting of the Association for Computational Linguistics},
 journal = {ArXiv},
 title = {AgentTuning: Enabling Generalized Agent Abilities for LLMs},
 volume = {abs/2310.12823},
 year = {2023},
 url = {https://arxiv.org/abs/2310.12823}
}

@Article{Chen2023FireActTL,
 author = {Baian Chen and Chang Shu and Ehsan Shareghi and Nigel Collier and Karthik Narasimhan and Shunyu Yao},
 booktitle = {arXiv.org},
 journal = {ArXiv},
 title = {FireAct: Toward Language Agent Fine-tuning},
 volume = {abs/2310.05915},
 year = {2023},
 url = {https://arxiv.org/abs/2310.05915}
}

@Article{Murty2024BAGELBA,
 author = {Shikhar Murty and Christopher D. Manning and Peter Shaw and Mandar Joshi and Kenton Lee},
 booktitle = {International Conference on Machine Learning},
 journal = {ArXiv},
 title = {BAGEL: Bootstrapping Agents by Guiding Exploration with Language},
 volume = {abs/2403.08140},
 year = {2024},
 url = {https://arxiv.org/abs/2403.08140}
}

@Article{Gunasekar2023TextbooksAA,
 author = {Suriya Gunasekar and Yi Zhang and J. Aneja and C. C. T. Mendes and A. Giorno and Sivakanth Gopi and Mojan Javaheripi and Piero Kauffmann and Gustavo de Rosa and Olli Saarikivi and A. Salim and S. Shah and Harkirat Singh Behl and Xin Wang and Sébastien Bubeck and Ronen Eldan and A. Kalai and Y. Lee and Yuan-Fang Li},
 booktitle = {arXiv.org},
 journal = {ArXiv},
 title = {Textbooks Are All You Need},
 volume = {abs/2306.11644},
 year = {2023},
 url = {https://arxiv.org/abs/2306.11644}
}

@Article{Ding2023EnhancingCL,
 author = {Ning Ding and Yulin Chen and Bokai Xu and Yujia Qin and Zhi Zheng and Shengding Hu and Zhiyuan Liu and Maosong Sun and Bowen Zhou},
 booktitle = {Conference on Empirical Methods in Natural Language Processing},
 journal = {ArXiv},
 title = {Enhancing Chat Language Models by Scaling High-quality Instructional Conversations},
 volume = {abs/2305.14233},
 year = {2023},
 url = {https://arxiv.org/abs/2305.14233}
}

@Article{Muennighoff2023ScalingDL,
 author = {Niklas Muennighoff and Alexander M. Rush and B. Barak and Teven Le Scao and Aleksandra Piktus and Nouamane Tazi and Sampo Pyysalo and Thomas Wolf and Colin Raffel},
 booktitle = {Neural Information Processing Systems},
 journal = {ArXiv},
 title = {Scaling Data-Constrained Language Models},
 volume = {abs/2305.16264},
 year = {2023},
 url = {https://arxiv.org/abs/2305.16264}
}

@Article{Albalak2024ASO,
 author = {Alon Albalak and Yanai Elazar and Sang Michael Xie and Shayne Longpre and Nathan Lambert and Xinyi Wang and Niklas Muennighoff and Bairu Hou and Liangming Pan and Haewon Jeong and Colin Raffel and Shiyu Chang and Tatsunori Hashimoto and W. Wang},
 booktitle = {Trans. Mach. Learn. Res.},
 journal = {ArXiv},
 title = {A Survey on Data Selection for Language Models},
 volume = {abs/2402.16827},
 year = {2024},
 url = {https://arxiv.org/abs/2402.16827}
}

@Article{Chiang2023CanLL,
 author = {Cheng-Han Chiang and Hung-yi Lee},
 booktitle = {Annual Meeting of the Association for Computational Linguistics},
 pages = {15607-15631},
 title = {Can Large Language Models Be an Alternative to Human Evaluations?},
 year = {2023},
 url = {https://arxiv.org/abs/2305.01937}
}

@Article{Kim2023PrometheusIF,
 author = {Seungone Kim and Jamin Shin and Yejin Cho and Joel Jang and S. Longpre and Hwaran Lee and Sangdoo Yun and Seongjin Shin and Sungdong Kim and James Thorne and Minjoon Seo},
 booktitle = {International Conference on Learning Representations},
 journal = {ArXiv},
 title = {Prometheus: Inducing Fine-grained Evaluation Capability in Language Models},
 volume = {abs/2310.08491},
 year = {2023},
 url = {https://arxiv.org/abs/2310.08491}
}

@Article{Yang2024Qwen25TR,
 author = {Qwen An Yang and Baosong Yang and Beichen Zhang and Binyuan Hui and Bo Zheng and Bowen Yu and Chengyuan Li and Dayiheng Liu and Fei Huang and Guanting Dong and Haoran Wei and Huan Lin and Jian Yang and Jianhong Tu and Jianwei Zhang and Jianxin Yang and Jiaxin Yang and Jingren Zhou and Junyang Lin and Kai Dang and Keming Lu and Keqin Bao and Kexin Yang and Le Yu and Mei Li and Mingfeng Xue and Pei Zhang and Qin Zhu and Rui Men and Runji Lin and Tianhao Li and Tingyu Xia and Xingzhang Ren and Xuancheng Ren and Yang Fan and Yang Su and Yi-Chao Zhang and Yunyang Wan and Yuqi Liu and Zeyu Cui and Zhenru Zhang and Zihan Qiu and Shanghaoran Quan and Zekun Wang},
 booktitle = {arXiv.org},
 journal = {ArXiv},
 title = {Qwen2.5 Technical Report},
 volume = {abs/2412.15115},
 year = {2024},
 url = {https://arxiv.org/abs/2412.15115}
}

@Article{Schulman2017ProximalPO,
 author = {John Schulman and Filip Wolski and Prafulla Dhariwal and Alec Radford and Oleg Klimov},
 booktitle = {arXiv.org},
 journal = {ArXiv},
 title = {Proximal Policy Optimization Algorithms},
 volume = {abs/1707.06347},
 year = {2017},
 url = {https://arxiv.org/abs/1707.06347}
}

@Article{DeepSeek-AI2025DeepSeekR1IR,
 author = {DeepSeek-AI and Daya Guo and Dejian Yang and Haowei Zhang and Jun-Mei Song and Ruoyu Zhang and R. Xu and Qihao Zhu and Shirong Ma and Peiyi Wang and Xiaoling Bi and Xiaokang Zhang and Xingkai Yu and Yu Wu and Z. F. Wu and Zhibin Gou and Zhihong Shao and Zhuoshu Li and Ziyi Gao and A. Liu and Bing Xue and Bing-Li Wang and Bochao Wu and B. Feng and Chengda Lu and Chenggang Zhao and C. Deng and Chenyu Zhang and C. Ruan and Damai Dai and Deli Chen and Dong-Li Ji and Erhang Li and Fangyun Lin and Fucong Dai and Fuli Luo and Guangbo Hao and Guanting Chen and Guowei Li and H. Zhang and Han Bao and Hanwei Xu and Haocheng Wang and Honghui Ding and Huajian Xin and Huazuo Gao and Hui Qu and Hui Li and Jianzhong Guo and Jiashi Li and Jiawei Wang and JingChang Chen and Jingyang Yuan and Junjie Qiu and Junlong Li and J. Cai and J. Ni and Jian Liang and Jin Chen and Kai Dong and Kai Hu and Kaige Gao and Kang Guan and Kexin Huang and K. Yu and Lean Wang and Lecong Zhang and Liang Zhao and Litong Wang and Liyue Zhang and Lei Xu and Leyi Xia and Mingchuan Zhang and Minghua Zhang and M. Tang and Meng Li and Miaojun Wang and Mingming Li and Ning Tian and Panpan Huang and Peng Zhang and Qiancheng Wang and Qinyu Chen and Qiushi Du and Ruiqi Ge and Ruisong Zhang and Ruizhe Pan and Runji Wang and R. J. Chen and R. Jin and Ruyi Chen and Shanghao Lu and Shangyan Zhou and Shanhuang Chen and Shengfeng Ye and Shiyu Wang and Shuiping Yu and Shunfeng Zhou and Shuting Pan and S. Li and Shuang Zhou and Shao-Kang Wu and Tao Yun and Tian Pei and T. Sun and T. Wang and Wangding Zeng and Wanjia Zhao and Wen Liu and W. Liang and Wenjun Gao and Wen-Xia Yu and Wentao Zhang and W. Xiao and Wei An and Xiaodong Liu and Xiaohan Wang and Xiaokang Chen and X. Nie and Xin Cheng and Xin Liu and Xin Xie and Xingchao Liu and Xinyu Yang and Xinyuan Li and Xuecheng Su and Xuheng Lin and X. Q. Li and Xiangyu Jin and Xi-Cheng Shen and Xiaosha Chen and Xiaowen Sun and Xiaoxiang Wang and Xinnan Song and Xinyi Zhou and Xianzu Wang and Xinxia Shan and Y. K. Li and Y. Q. Wang and Y. X. Wei and Yang Zhang and Yanhong Xu and Yao Li and Yao Zhao and Yaofeng Sun and Yaohui Wang and Yi Yu and Yichao Zhang and Yifan Shi and Yi Xiong and Ying He and Y. Piao and Yisong Wang and Yixuan Tan and Yiyang Ma and Yiyuan Liu and Yongqiang Guo and Y. Ou and Yuduan Wang and Yue Gong and Yu-Jing Zou and Yujia He and Yunfan Xiong and Yu-Wei Luo and Yu-mei You and Yuxuan Liu and Yuyang Zhou and Y. X. Zhu and Yanping Huang and Yao Li and Yi Zheng and Yuchen Zhu and Yunxiang Ma and Ying Tang and Y. Zha and Yuting Yan and Z. Ren and Z. Ren and Zhangli Sha and Zhe Fu and Zhean Xu and Zhenda Xie and Zhen-guo Zhang and Zhewen Hao and Zhicheng Ma and Zhigang Yan and Zhiyu Wu and Zihui Gu and Zijia Zhu and Zijun Liu and Zi-An Li and Ziwei Xie and Ziyang Song and Zizheng Pan and Zhen Huang and Zhipeng Xu and Zhongyu Zhang and Zhen Zhang},
 booktitle = {Nature},
 journal = {Nature},
 pages = {633 - 638},
 title = {DeepSeek-R1 incentivizes reasoning in LLMs through reinforcement learning},
 volume = {645},
 year = {2025},
 url = {https://arxiv.org/abs/2501.12948}
}

@Article{Christiano2017DeepRL,
 author = {P. Christiano and Jan Leike and Tom B. Brown and Miljan Martic and S. Legg and Dario Amodei},
 booktitle = {Neural Information Processing Systems},
 journal = {ArXiv},
 title = {Deep Reinforcement Learning from Human Preferences},
 volume = {abs/1706.03741},
 year = {2017},
 url = {https://arxiv.org/abs/1706.03741}
}

@Article{Snow2008CheapAF,
 author = {R. Snow and Brendan T. O'Connor and Dan Jurafsky and A. Ng},
 booktitle = {Conference on Empirical Methods in Natural Language Processing},
 pages = {254-263},
 title = {Cheap and Fast – But is it Good? Evaluating Non-Expert Annotations for Natural Language Tasks},
 year = {2008},
 url = {https://doi.org/10.3115/1613715.1613751}
}

@Article{Gururangan2018AnnotationAI,
 author = {Suchin Gururangan and Swabha Swayamdipta and Omer Levy and Roy Schwartz and Samuel R. Bowman and Noah A. Smith},
 booktitle = {North American Chapter of the Association for Computational Linguistics},
 journal = {ArXiv},
 title = {Annotation Artifacts in Natural Language Inference Data},
 volume = {abs/1803.02324},
 year = {2018},
 url = {https://arxiv.org/abs/1803.02324}
}

@Article{Madaan2023SelfRefineIR,
 author = {Aman Madaan and Niket Tandon and Prakhar Gupta and Skyler Hallinan and Luyu Gao and Sarah Wiegreffe and Uri Alon and Nouha Dziri and Shrimai Prabhumoye and Yiming Yang and S. Welleck and Bodhisattwa Prasad Majumder and Shashank Gupta and A. Yazdanbakhsh and Peter Clark},
 booktitle = {Neural Information Processing Systems},
 journal = {ArXiv},
 title = {Self-Refine: Iterative Refinement with Self-Feedback},
 volume = {abs/2303.17651},
 year = {2023},
 url = {https://arxiv.org/abs/2303.17651}
}

@Article{Shinn2023ReflexionLA,
 author = {Noah Shinn and Federico Cassano and Beck Labash and A. Gopinath and Karthik Narasimhan and Shunyu Yao},
 booktitle = {Neural Information Processing Systems},
 journal = {Advances in Neural Information Processing Systems 36},
 title = {Reflexion: language agents with verbal reinforcement learning},
 year = {2023},
 url = {https://arxiv.org/abs/2303.11366}
}

@Article{Snell2024ScalingLT,
 author = {C. Snell and Jaehoon Lee and Kelvin Xu and Aviral Kumar},
 booktitle = {arXiv.org},
 journal = {ArXiv},
 title = {Scaling LLM Test-Time Compute Optimally can be More Effective than Scaling Model Parameters},
 volume = {abs/2408.03314},
 year = {2024},
 url = {https://arxiv.org/abs/2408.03314}
}

@Article{Gulcehre2023ReinforcedS,
 author = {Caglar Gulcehre and T. Paine and S. Srinivasan and Ksenia Konyushkova and L. Weerts and Abhishek Sharma and Aditya Siddhant and Alexa Ahern and Miaosen Wang and Chenjie Gu and Wolfgang Macherey and A. Doucet and Orhan Firat and Nando de Freitas},
 booktitle = {arXiv.org},
 journal = {ArXiv},
 title = {Reinforced Self-Training (ReST) for Language Modeling},
 volume = {abs/2308.08998},
 year = {2023},
 url = {https://arxiv.org/abs/2308.08998}
}

@Article{Tunstall2023ZephyrDD,
 author = {Lewis Tunstall and E. Beeching and Nathan Lambert and Nazneen Rajani and Kashif Rasul and Younes Belkada and Shengyi Huang and L. V. Werra and Clémentine Fourrier and Nathan Habib and Nathan Sarrazin and Omar Sanseviero and Alexander M. Rush and Thomas Wolf},
 booktitle = {arXiv.org},
 journal = {ArXiv},
 title = {Zephyr: Direct Distillation of LM Alignment},
 volume = {abs/2310.16944},
 year = {2023},
 url = {https://arxiv.org/abs/2310.16944}
}

@Article{Wei2022ChainOT,
 author = {Jason Wei and Xuezhi Wang and Dale Schuurmans and Maarten Bosma and Ed H. Chi and F. Xia and Quoc Le and Denny Zhou},
 booktitle = {Neural Information Processing Systems},
 journal = {ArXiv},
 title = {Chain of Thought Prompting Elicits Reasoning in Large Language Models},
 volume = {abs/2201.11903},
 year = {2022},
 url = {https://arxiv.org/abs/2201.11903}
}

@Article{Yao2023TreeOT,
 author = {Shunyu Yao and Dian Yu and Jeffrey Zhao and Izhak Shafran and T. Griffiths and Yuan Cao and Karthik Narasimhan},
 booktitle = {Neural Information Processing Systems},
 journal = {ArXiv},
 title = {Tree of Thoughts: Deliberate Problem Solving with Large Language Models},
 volume = {abs/2305.10601},
 year = {2023},
 url = {https://arxiv.org/abs/2305.10601}
}

@Book{Kwon2023EfficientMM,
 author = {Woosuk Kwon and Zhuohan Li and Siyuan Zhuang and Ying Sheng and Lianmin Zheng and Cody Hao Yu and Joseph E. Gonzalez and Haotong Zhang and Ion Stoica},
 booktitle = {Symposium on Operating Systems Principles},
 journal = {Proceedings of the 29th Symposium on Operating Systems Principles},
 title = {Efficient Memory Management for Large Language Model Serving with PagedAttention},
 year = {2023},
 url = {https://arxiv.org/abs/2309.06180}
}

@Article{Dao2022FlashAttentionFA,
 author = {Tri Dao and Daniel Y. Fu and Stefano Ermon and A. Rudra and Christopher R'e},
 booktitle = {Neural Information Processing Systems},
 journal = {ArXiv},
 title = {FlashAttention: Fast and Memory-Efficient Exact Attention with IO-Awareness},
 volume = {abs/2205.14135},
 year = {2022},
 url = {https://arxiv.org/abs/2205.14135}
}

@Article{Zhao2023PyTorchFE,
 author = {Yanli Zhao and A. Gu and R. Varma and Liangchen Luo and Chien-chin Huang and Min Xu and Less Wright and Hamid Shojanazeri and Myle Ott and Sam Shleifer and Alban Desmaison and Can Balioglu and Bernard Nguyen and Geeta Chauhan and Yuchen Hao and Shen Li},
 booktitle = {Proceedings of the VLDB Endowment},
 journal = {Proc. VLDB Endow.},
 pages = {3848-3860},
 title = {PyTorch FSDP: Experiences on Scaling Fully Sharded Data Parallel},
 volume = {16},
 year = {2023},
 url = {https://arxiv.org/abs/2304.11277}
}

@Article{Guo2026OpAgentOA,
 author = {Yuyu Guo and Wenjie Yang and Siyuan Yang and Ziyang Liu and Cheng Chen and Yuan Wei and Yun Hu and Yangru Huang and Guo Hao and Dongsheng Yuan and Jianming Wang and Xin Chen and Hang Yu and Lei Lei and Peng Di},
 booktitle = {arXiv.org},
 journal = {ArXiv},
 title = {OpAgent: Operator Agent for Web Navigation},
 volume = {abs/2602.13559},
 year = {2026},
 url = {https://arxiv.org/abs/2602.13559}
}

@Article{Gandhi2025GoBrowseTW,
 author = {Apurva Gandhi and Graham Neubig},
 booktitle = {arXiv.org},
 journal = {ArXiv},
 title = {Go-Browse: Training Web Agents with Structured Exploration},
 volume = {abs/2506.03533},
 year = {2025},
 url = {https://arxiv.org/abs/2506.03533}
}

@Article{Wei2025WebAgentR1TW,
 author = {Zhepei Wei and Wenlin Yao and Yao Liu and Weizhi Zhang and Qin Lu and Liang Qiu and Changlong Yu and Puyang Xu and Chao Zhang and Bing Yin and Hyokun Yun and Lihong Li},
 booktitle = {Conference on Empirical Methods in Natural Language Processing},
 pages = {7909-7928},
 title = {WebAgent-R1: Training Web Agents via End-to-End Multi-Turn Reinforcement Learning},
 year = {2025},
 url = {https://arxiv.org/abs/2505.16421}
}

@Article{Andrade2025LetsTI,
 author = {Moises Andrade and Joonhyuk Cha and Brandon Ho and V. Srihari and Karmesh Yadav and Z. Kira},
 booktitle = {arXiv.org},
 journal = {ArXiv},
 title = {Let's Think in Two Steps: Mitigating Agreement Bias in MLLMs with Self-Grounded Verification},
 volume = {abs/2507.11662},
 year = {2025},
 url = {https://arxiv.org/abs/2507.11662}
}

@Article{Shen2025ThinkingVD,
 author = {Junhong Shen and Hao Bai and Lunjun Zhang and Yifei Zhou and Amrith Rajagopal Setlur and Shengbang Tong and Diego Caples and Nan Jiang and Tong Zhang and Ameet Talwalkar and Aviral Kumar},
 booktitle = {arXiv.org},
 journal = {ArXiv},
 title = {Thinking vs. Doing: Agents that Reason by Scaling Test-Time Interaction},
 volume = {abs/2506.07976},
 year = {2025},
 url = {https://arxiv.org/abs/2506.07976}
}

@Article{Koh2024TreeSF,
 author = {Jing Yu Koh and S. McAleer and Daniel Fried and Ruslan Salakhutdinov},
 booktitle = {Trans. Mach. Learn. Res.},
 journal = {ArXiv},
 title = {Tree Search for Language Model Agents},
 volume = {abs/2407.01476},
 year = {2024},
 url = {https://arxiv.org/abs/2407.01476}
}

@Article{Sarch2025GroundedRL,
 author = {Gabriel Sarch and Snigdha Saha and Naitik Khandelwal and Ayush Jain and Michael J. Tarr and Aviral Kumar and Katerina Fragkiadaki},
 booktitle = {arXiv.org},
 journal = {ArXiv},
 title = {Grounded Reinforcement Learning for Visual Reasoning},
 volume = {abs/2505.23678},
 year = {2025},
 url = {https://arxiv.org/abs/2505.23678}
}

@Article{Wang2024GUIAW,
 author = {Shuai Wang and Weiwen Liu and Jingxuan Chen and Weinan Gan and Xingshan Zeng and Shuai Yu and Xinlong Hao and Kun Shao and Yasheng Wang and Ruiming Tang},
 booktitle = {arXiv.org},
 journal = {ArXiv},
 title = {GUI Agents with Foundation Models: A Comprehensive Survey},
 volume = {abs/2411.04890},
 year = {2024},
 url = {https://arxiv.org/abs/2411.04890}
}

@Article{Penaloza2026PrivilegedID,
 author = {Emiliano Penaloza and Dheeraj Vattikonda and Nicolas Gontier and Alexandre Lacoste and Laurent Charlin and Massimo Caccia},
 booktitle = {arXiv.org},
 journal = {ArXiv},
 title = {Privileged Information Distillation for Language Models},
 volume = {abs/2602.04942},
 year = {2026},
 url = {https://arxiv.org/abs/2602.04942}
}
